\documentclass[preprint,12pt,3p]{elsarticle}
\usepackage{microtype}
\usepackage{floatpag}
\usepackage{graphicx}
\usepackage{subfiles}
\usepackage{subcaption}
\usepackage{textcomp}
\usepackage{amsmath,amssymb}
\DeclareMathOperator{\E}{\mathbb{E}}

\usepackage{booktabs} 
\newcommand{\vect}[1]{\boldsymbol{#1}}
\usepackage{hyperref}
\usepackage{pbox}
\usepackage{algorithm}
\usepackage{algpseudocode}
\usepackage{setspace}
\newcommand{\pluseq}{\mathrel{+}=}

\journal{NeuroImage}

\begin{document}

\begin{frontmatter}

\title{Monotonic Gaussian Process for Spatio-Temporal Disease Progression Modeling in Brain Imaging Data}

\author[label1]{Cl\'ement Abi Nader\corref{cor1}}
\address[label1]{Universit\'e C\^ote d'Azur, Inria Sophia Antipolis, Epione Research Project, France.}
\cortext[cor1]{Corresponding author at: Epione Research Project, INRIA Sophia-Antipolis, 2004, route des Lucioles, 06902 Sophia-Antipolis, France, clement.abi-nader@inria.fr.}


\ead{clement.abi-nader@inria.fr}

\author[label1]{Nicholas Ayache}
\ead{nicholas.ayache@inria.fr}

\author[label2]{Philippe Robert}
\address[label2]{Universit\'e C\^ote d'Azur, CoBTeK lab, MNC3 program, France.}
\ead{probert@unice.fr}

\author[label1]{Marco Lorenzi}
\ead{marco.lorenzi@inria.fr}

\author{for the Alzheimer's Disease Neuroimaging Initiative$^{**}$}
\cortext[cor2]{Data used in preparation of this article were obtained from the Alzheimer\textquotesingle s Disease Neuroimaging Initiative (ADNI) database (adni.loni.usc.edu). As such, the investigators within the ADNI contributed to the design and implementation of ADNI and/or provided data but did not participate in analysis or writing of this report. A complete listing of ADNI investigators can be found at: http://adni.loni.usc.edu/wp-content/uploads/how\textunderscore to\textunderscore apply/ADNI\textunderscore Acknowledgement\textunderscore List.pdf.}

\begin{abstract}
We introduce a probabilistic generative model for disentangling spatio-temporal disease trajectories from collections of high-dimensional brain images. The model is based on spatio-temporal matrix factorization, where inference on the sources is constrained by anatomically plausible statistical priors. To model realistic trajectories, the temporal sources are defined as monotonic and time-reparameterized Gaussian Processes. To account for the non-stationarity of brain images, we model the spatial sources as sparse codes convolved at multiple scales. The method was tested on synthetic data favourably comparing with standard blind source separation approaches. The application on large-scale imaging data from a clinical study allows to disentangle differential temporal progression patterns mapping brain regions key to neurodegeneration, while revealing a disease-specific time scale associated to the clinical diagnosis. 
\end{abstract}


\end{frontmatter}



\section{Introduction}
\label{sec:intro}
\documentclass[main.tex]{subfiles}
Neurodegenerative disorders such as Alzheimer's disease (AD) are characterized by morphological and molecular changes of the brain, ultimately leading to cognitive and behavioral decline. Clinicians suggested hypothetical models of the disease evolution, showing how different types of biomarkers interact and lead to the final dementia stage \cite{ref_clifford}. In the past years, efforts have been made in order to collect large databases of imaging and clinical measures, hoping to obtain more insights about the disease progression through data-driven models describing the trajectory of the disease over time. This kind of models are of critical importance for understanding the pathological progression in large scale data, and would represent a valuable reference for improving the individual diagnosis. 
\newline
\newline
Current clinical trials in AD are based on longitudinal monitoring of biomarkers. Disease progression modelling aims at providing an interpretable way of modelling the evolution of biomarkers according to an estimated history of the pathology, as proposed for example in \cite{ref_donohue}, \cite{ref_fonteijn}, \cite{ref_jedynak}, \cite{ref_lorenzi}, and \cite{ref_young_2014}. Therefore, disease progression models are promising methods for automatically staging patients, and quantifying their progression with respect to the underlying model of the pathology. These approaches entail a great potential for automatic stratification of individuals based on their estimated stage and progression speed, and for assessment of efficacy of disease modifying drugs. Within this context, we propose a spatio-temporal generative model of disease progression, aimed at disentangling and quantifying the independent dynamics of changes observed in datasets of multi-modal data. With this term we indicate data acquired via different imaging modalities such as Magnetic Resonance Imaging (MRI) or Positron-Emission Tomography (PET), as well as non-imaging data such as clinical scores assessed by physicians. Moreover, we aim at automatically inferring the disease severity of a patient with respect to the estimated trajectory. Defining such a disease progression model raises a number of methodological challenges. 
\newline
\newline
AD spreads over decades with a temporal mismatch between the onset of the disease and the moment where the clinical symptoms appear.  Either age of diagnosis, or the chronological age, are therefore not suitable as a temporal reference to describe the disease progression in time. Moreover, as the follow-up of patients doesn't exceed a few years, the development of a model of long-term pathological changes requires to integrate cross-sectional data from different individuals, in order to consider a longer period of time. In virtue of the lack of a well defined temporal reference, observations from different individuals are characterized by large and unknown variability in the onset and speed of the disease. It is therefore necessary to account for a time-reparameterization function, mapping each individuals' observations to a common temporal axis associated to the absolute disease trajectory \cite{ref_jedynak,ref_schiratti}. This would allow to estimate an absolute time-reference related to the natural history of the pathology. 
\newline
\newline
The analysis of MRI and PET data, requires to account for spatio-temporally correlated features (voxels, i.e. volumetric pixels) defined over arrays of more than a million entries. The development of inference schemes jointly considering these correlation properties thus raises  scalability issues, especially when accounting for the non-stationarity of the image signal. Furthermore, the brain regions involved in AD exhibit various dynamics in time, and evolve at different speed \cite{ref_whitwell}. From a modeling perspective, accounting for differential trajectories over space and time raises the problem of source identification and separation. This issue has been widely addressed in neuroimaging via Independent Component Analysis (ICA) \cite{ref_ica}, especially on functional MRI (fMRI) data \cite{ref_fmri}. Nevertheless, while fMRI time-series are usually defined over a few hundreds of time points acquired per subject, our problem consists in jointly analyzing short-term and cross-sectional data observations with respect to an unknown time-line. This problem cannot be tackled with standard ICA, as time is generally an independent variable on which inference is not required. Moreover, ICA retrieves spatial sources based on the assumption of statistical independence. This assumption does not necessarily lead to clinically interpretable findings. Indeed, dependency across temporal patterns can be still highly relevant to the pathology, for example when modeling temporal delay across similar sources.
\newline
\newline
The problem of providing a realistic description of the biological processes is critical when analyzing biomedical data, such as medical images. For example, to describe a plausible evolution of AD from normal to pathological stages, smoothness and monotonicity are commonly assumed for the temporal sources. It is also necessary to account for the non-stationarity of changes affecting the brain from global to localized spatio-temporal processes. As a result, spatial sources need to account for different resolutions at which these changes take place. While several multi-scale analysis approaches have been proposed to model spatio-temporal signals \cite{ref_mallat,ref_bullmore,ref_hackmack}, extending this type of methods to the high-dimension of medical images is generally not trivial due to scalability issues. Finally, the noisy nature of medical images, along with the large signal variability across observations, requires a modeling framework robust to bias and noise. 
\newline
\newline
In this work, we propose to jointly address these issues within a Bayesian framework for the spatio-temporal analysis of large-scale collections of multi-modal brain data. We show that this framework allows us to naturally encode plausibility constraints through clinically-inspired priors, while accounting for the uncertainty of the temporal profiles and brain structures we wish to estimate. Similarly to the ICA setting, we formulate the problem of trajectory modeling through matrix factorization across temporal and spatial sources. This is done for each modality by inferring their specific spatio-temporal sources. To promote smoothness in time and avoid any unnecessary hypothesis on the temporal trajectories, we rely on non-parametric modeling based on Gaussian Process (GP). We account for a plausible evolution from healthy to pathological stages thanks to a monotonicity constraint applied on the GP. Moreover, individuals' observations are temporally re-aligned on a common scale via a time-warping function. In case of imaging data, to model the non-stationarity of the spatial signal, the spatial sources are defined as sparse activation maps convolved at different scales. We show that our framework can be efficiently optimized through stochastic variational inference, allowing to exploit automatic differentiation and GPU support to speed up computations. 
\newline
\newline
The paper is organized as follows: Section \ref{sec:related_work} analyzes related work on spatio-temporal modeling of neurodegeneration, while Section \ref{sec:method} details our method. In Section \ref{sec:results} we present experiments on synthetic data in which we compare our model to standard blind source separation approaches. We finally provide a demonstration of our method on the modeling of imaging data from a large scale clinical study. 
Prospects for future work and conclusions are drawn in section \ref{sec:discussion}. Derivations that we could not fit in the paper are detailed in the Appendices.

\section{Related Work in Neurodegeneration Modeling}
\label{sec:related_work}
\documentclass[main.tex]{subfiles}
To deal with the uncertainty of the time-line of neurodegenerative pathologies, the concept of time-reparameterization of imaging-derived features has been used in several works. The underlying principle consists in estimating an absolute time-scale of disease progression by temporally re-aligning data from different subjects.  For instance, in \cite{ref_young} the time-evolution was approximated as a sequence of events which need to be re-ordered for each patient. 
This approach thus considers the evolution of neurodegenerative diseases as a collection of transitions between discrete stages. This hypothesis is however limiting, as it doesn't reflect the continuity of changes affecting the brain along the course of the pathology.
\newline
\newline
To address this limitation, we rely on a continuous parameterization of the time-axis as in \cite{ref_lorenzi,ref_donohue}. In particular, individuals' observations are time-realigned on a common temporal scale via a time-warping function. Using a set of relevant scalar biomarkers, this kind of approach allows to learn a time-scale describing the pathology evolution, and to estimate a data-driven time-line markedly correlated with the decline of cognitive abilities. Similarly, in \cite{ref_bilgel} a disease progression score was estimated using biomarkers from molecular imaging. These methods are however based on the analysis of low-dimensional measures, such as collections of clinical variables. Therefore, they do not allow to scale to the high dimension of multi-modal medical images. Our work tackles this shortcoming thanks to a scalable inference scheme based on stochastic variational inference.
\newline
\newline
Concerning the spatio-temporal representation of neurodegeneration, a mixed-effect model was proposed by \cite{ref_koval} to learn an average spatio-temporal trajectory of brain evolution on cortical thickness data. The fixed-effect describes the average trajectory, while random effects are estimated through individual spatio-temporal warping functions, modeling how each subject differs from the global progression. Still, the extension of this approach to image volumes  raises scalability issues. It has also to be noted that, to allow computational tractability, the brain evolution was assumed to be stationary both in space and time, thus limiting the ability of the model to disentangle the multiple dynamics of the brain structures involved in AD. 
\newline
\newline
An attempt to source separation is proposed in \cite{ref_marinescu}, through the decomposition of cortical thickness measurements as a mixture of spatio-temporal processes. This is performed by associating to each cortical vertex a temporal progression modeled by a sigmoid function, which may be however too simplistic to describe the progression of AD temporal processes. We propose to overcome this issue by non-parametric modeling of the temporal sources through GPs. Moreover, the model in \cite{ref_marinescu} is lacking of an explicit vertex-wise correlation model, as it only assumes correlation between clustering parameters at the resolution of the mesh graph. For this reason, it may still be sensitive to spatial variation at different scales and noise. We address this problem by modeling the spatial sources through convolution of sparse maps at multiple resolutions, allowing to deal with signal non-stationarity and robustness to noise.

\section{Methods}
\label{sec:method}
\documentclass[main.tex]{subfiles}
In the following sections a matrix will be denoted by an uppercase letter $\vect{X}$, its n-th row will be given by $\vect{X}_{n:}$ and its n-th column by $\vect{X}_{:n}$. A column vector will be denoted by a lowercase letter $\vect{x}$. Subscript indices will be used to index the elements of matrices, vectors or sets of scalars. Superscipt indices will allow to index the blocks of block diagonal matrices. 
\subsection{Individual time-shift}
\label{ssec:time_shift}
To account for the uncertainty of the time-line of individual measurements, we assume that the observations are defined with respect to an absolute temporal reference $\tau$. This is performed through a time-warping function $t_{p} = \vect{f}_{p}(\tau)$, that models the individual time-reparameterization. We choose an additive parameterization such that:
\begin{equation}
  \vect{f}_{p}(\tau) = \tau + \delta_{p}.   
\end{equation}
Within this setting the individual time-shift $\delta_{p}$ encodes the temporal position of subject $p$, which in our application can be interpreted as the disease stage of subject $p$ with respect to the long-term disease trajectory. We denote by $\vect{\delta} = \{\delta_p\}_{p=0}^{P}$ the set of time-shift parameters.

\subsection{Data modeling}
\label{ssec:data_modelling}
We represent the spatio-temporal data $\vect{D}$ by a block diagonal matrix in which we differentiate two main blocks $\vect{Y}$ and $\vect{V}$ as illustrated in Figure \ref{fig:method_overview}. Each sub-block $\vect{Y}^{m}$ is a matrix containing the data represented by one of the $M$ imaging modalities we wish to consider. These matrices have dimensions $P \times F_{m}$, where $P$ denotes the number of subjects and $F_{m}$ the number of imaging features for modality $m$, which in our case is the number of voxels. The matrix $\vect{V}$ accounts for non-imaging or scalar data such as clinical scores and has dimensions $P \times C$, where $C$ is the number of scalar features considered. We postulate a generative model and decompose the data as shown in Figure \ref{fig:method_overview}.
\begin{figure}[h!]
\centering
\includegraphics[width=1.\textwidth]{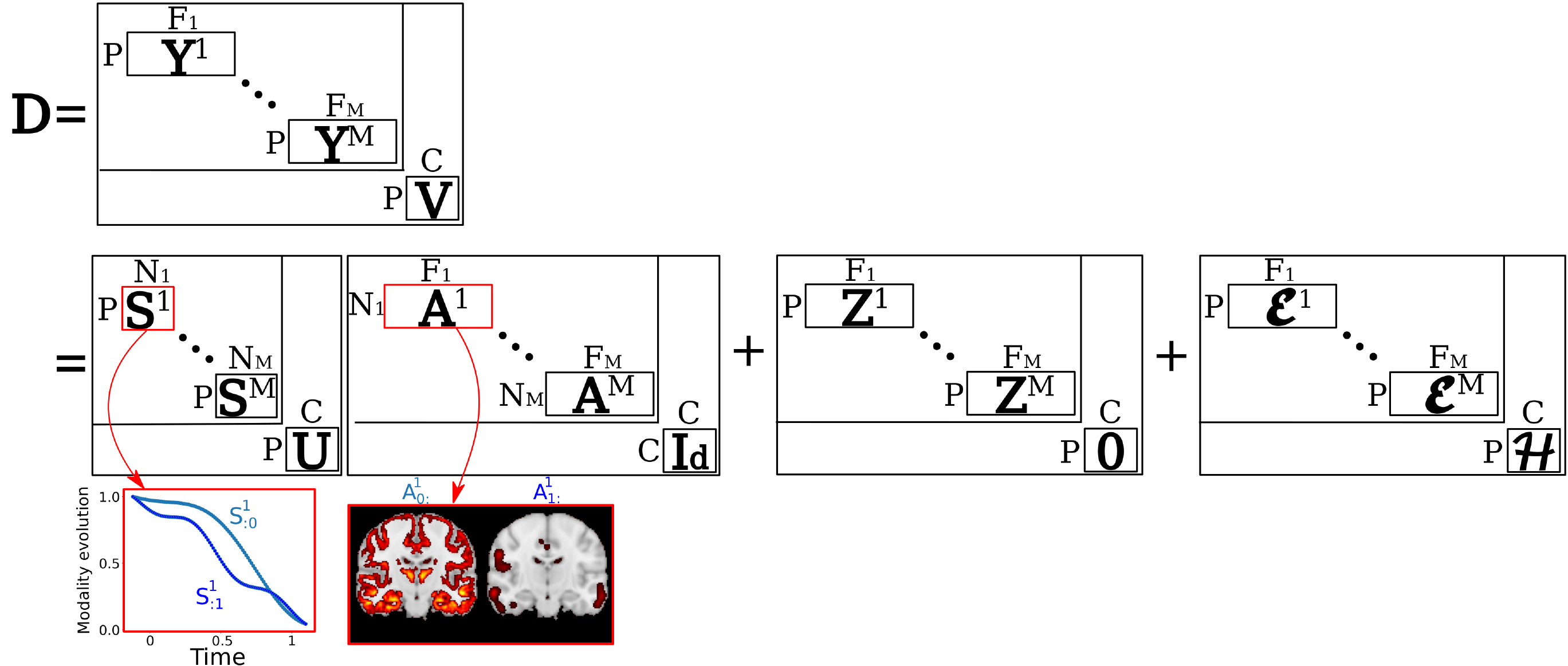}
\caption{Spatio-temporal decomposition of each data block. A data matrix composed by M imaging modalities is decomposed as the product of monotonic temporal sources $\vect{S}^{m}$ and corresponding activation maps $\vect{A}^{m}$. Monotonic sources are also used to model the scalar biomarkers $\vect{V}$, while we assume additive constant terms $\vect{Z}^{m}$, and noise $\vect{\mathcal{E}}^{m}$.}
\label{fig:method_overview}
\end{figure}
For each sub-block $\vect{Y}^{m}$, the data is factorized in a set of $N_{m}$ spatio-temporal sources $\vect{Y}^{m}=\vect{S}^{m}\vect{A}^{m}$. The columns of the matrix $\vect{S}^{m}$ describe the non-linear temporal evolution of the corresponding spatial maps contained in the rows of $\vect{A}^{m}$. Therefore, their product represents the voxel-wise linear combination of the spatial maps modulated by the corresponding temporal sources. The subjects share the same set of temporal sources across $\vect{S}^1,..,\vect{S}^M$, as these sources describe the temporal evolution of the group-wise images through the regression problem specified in Figure \ref{fig:method_overview}. The data in matrix $\vect{V}$ is modelled by a matrix $\vect{U}$ whose columns depict the temporal trajectories of the different scalar scores. In the case of imaging data, we also consider a constant term modeling brain areas which don't exhibit any intensity changes over time. This is done by including constant matrix terms $\vect{Z}^{m}$ that we need to estimate. We assume for a given modality $m$ that the vectors $\vect{Z}^{m}_{p:}$ are common to  every subjects. Finally, for each modality $m$, scalar score $c$, and subject $p$, we assume Gaussian observational noise $\vect{\mathcal{E}}^{m}_{p:} \sim \mathcal{N}(\vect{0}, \sigma^{2}_{m}\vect{I})$, and $\vect{\mathcal{H}}_{p,c} \sim \mathcal{N}(0, \nu^{2}_{c})$ for respectively imaging and scalar information. \\

Therefore, if we consider the data from modality $m$ and scalar $c$ of patient $p$ observed at time $\vect{f}_{p}(\tau)$ we have:
\begin{align} 
\begin{split}
\label{eq:data_decomposition}
\displaystyle 
& \vect{Y}_{p:}^{m}(\vect{f}_p(\tau),\theta_{m}, \psi_{m}) = \vect{S}_{p:}^{m}(\vect{f}_{p}(\tau), \theta_{m})\vect{A}^{m}(\psi_{m}) + \vect{Z}_{p:}^{m} + \vect{\mathcal{E}}_{p:}^{m}, \\
& \vect{V}_{p,c}(\vect{f}_p(\tau),\theta_{c}) = \vect{U}_{p,c}(\vect{f}_{p}(\tau), \theta_{c}) + \vect{\mathcal{H}}_{p,c}.
\end{split}
\end{align}
We denote by $\theta_m$ and $\theta_c$ the temporal parameters related respectively to the modality m and scalar feature c, while $\psi_m$ represents the set of spatial parameters of modality m. We assume conditional independence across modalities and scalar scores given the time-shift information: 
\begin{align}
\begin{split}
\label{eq:to_optim1}
\displaystyle
p(\vect{Y}, \vect{V}|\vect{A},\vect{S},\vect{Z}, \vect{U}, \vect{\delta},\sigma, \nu) & =  \Big(\prod_{m} p(\vect{Y}^{m}|\vect{A}^{m},\vect{S}^{m}, \vect{Z}^{m}, \vect{\delta}, \sigma_{m}) \Big) \Big( \prod_{c} p(\vect{V}_{:c}|\vect{U}_{:c}, \vect{\delta}, \nu_{c}) \Big).
\end{split}
\end{align}
Relying on classical regression formulation, we assume exchangeability across subjects allowing us to derive the data likelihood for a given modality m. According to the generative model we can write:
\begin{align}
\begin{split}
\label{eq:to_optim2}
p(\vect{Y}^{m}|\vect{A}^{m},\vect{S}^{m}, \vect{Z}^{m}, \vect{\delta}, \sigma_{m}) =  \Big(\prod_{p} & \frac{1}{(2\pi \sigma_{m}^{2})^\frac{F_{m}}{2}} \exp(-\frac{1}{2\sigma_{m}^{2}} ||\vect{Y}_{p:}^{m}(\vect{f}_p(\tau),\theta_{m}, \psi_{m}) \\
& - \vect{S}_{p:}^{m}(\vect{f}_{p}(\tau), \theta_{m})\vect{A}^{m}(\psi_{m}) - \vect{Z}_{p:}^{m}||^{2}) \Big).
\end{split}
\end{align}
Naturally, a similar equation holds for $p(\vect{V}_{:c}|\vect{U}_{:c}, \vect{\delta}, \nu_{c})$.
\newline
\newline
Within a Bayesian modeling framework, we wish to maximize the marginal log-likelihood $\log(p(\vect{Y}, \vect{V}|\vect{Z}, \vect{\delta}, \sigma, \nu))$, to obtain posterior distributions for the spatio-temporal processes. Since the derivation of this quantity in a closed-form is not possible, we tackle this optimization problem through stochastic variational inference. Based on this formulation, in what follows we illustrate our model by detailing the variational approximations imposed on the spatio-temporal sources, along with the priors and constraints we impose to represent the data (Sections \ref{ssec:spatio_temporal_processes} and \ref{ssec:sparsity}). Finally, we detail the variational lower bound and optimization strategy in Section \ref{ssec:vif}. \\

For ease of notation we will drop the $m$ and $c$ indexes in Sections \ref{ssec:spatio_temporal_processes} and \ref{ssec:sparsity}. As a result the matrix $\vect{S}$ will indistinctly refer to either any $\vect{S}^{m}$ or $\vect{U}$, while matrix $\vect{A}$ will refer to any $\vect{A}^{m}$, and $\vect{Y}$ to any $\vect{Y}^{m}$. For a given modality $m$, the number of patients $P$ will be indexed by $p$, the number of sources $N^{m}$ or the number of scalar scores $C$ will be indexed by $n$, and finally $f$ will index the number of imaging features $F^{m}$.

\subsection{Spatio-temporal processes}
\label{ssec:spatio_temporal_processes}
\subsubsection{Temporal sources}
\label{sssec:temporal_sources}
In order to flexibly account for non-linear temporal patterns, the temporal sources are encoded in a matrix $\vect{S}$ in which each column $\vect{S}_{:n}$ is a GP representing the evolution of source $n$ and is independent from the other sources. To allow computational tractability within a variational setting, we rely on the GP approximation proposed in \cite{ref_gp}, through kernel approximation via random feature expansion \cite{rahimi}. Within this framework, a GP can be approximated as a Bayesian Neural Network with form: $\vect{S}_{:n}(\vect{t}) = \phi(\vect{t}(\vect{\omega}^{n})^{T})\vect{w}^{n}$. For example, in the case of the Radial Basis Function (RBF) covariance, $\vect{\omega}^{n}$ is a linear projection in the spectral domain. It is equipped with a Gaussian distributed prior $p(\vect{\omega}^{n}) \sim \mathcal{N}(\vect{0}, l_{n}\vect{I})$ with a zero-mean and a covariance  parameterized by a scalar $l_{n}$, acting as the length-scale parameter of the RBF covariance. The non-linear basis functions activation is defined by setting $\phi(\cdot) = (\cos(\cdot), \sin(\cdot))$, while the regression parameter $\vect{w}^{n}$ is given with a standard normal prior. The GP inference problem can be conveniently performed by estimating approximated variational distributions for all the $\vect{\omega}^{n}$ and $\vect{w}^{n}$ (Section \ref{ssec:vif}). We will respectively denote by $\vect{\Omega}$ and $\vect{W}$ the block diagonal matrices whose blocks are the $(\vect{\omega}^{n})^{T}$ and $\vect{w}^{n}$. Considering the $N$ temporal sources, we can write $p(\vect{\Omega}) = \prod_{n} p(\vect{\omega}^{n})$ and $p(\vect{W}) = \prod_{n} p(\vect{w}^{n})$.
\newline
\newline
We wish also to account for a steady evolution of the temporal processes, hence constraining the temporal sources to monotonicity. This is relevant in the medical case, where one would like to model the steady progression of a disease from normal to pathological stages. In our case, we want to constrain the space of the temporal sources to the set of solutions $\mathcal{C}_{n} = \{\vect{S}_{:n}(\vect{t})  \mid \vect{S}_{:n}'(\vect{t}) \geq 0 \quad \forall \ \vect{t}\}$.  This can be done done consistently within the regression setting of \cite{ref_vehtari}, and in particular with the GP random feature expansion framework as shown in \cite{ref_monotonicity}. In that work, the constraint is introduced as a second likelihood term on the temporal sources dynamics:
\begin{equation} 
p(\mathcal{C}|\vect{S}', \gamma) = \prod_{p,n} (1 + \exp(-\gamma \vect{S}_{p,n}'(\vect{t})))^{-1},
\end{equation}
where $\vect{S}'$ contains every derivatives $\vect{S}'_{:n}$, $\gamma$ controls the magnitude of the monotonicity constraint, and $\mathcal{C} = \bigcap_n \mathcal{C}_{n}$. According to  \cite{ref_monotonicity} this constraint can be specified through the parametric form for the derivative of each $\vect{S}_{:n}$:
\begin{equation}
    \vect{S}_{:n}'(t) = \frac{d\phi(\vect{t}(\vect{\omega}^{n})^{T})}{d\vect{t}}\vect{w}^{n}.
\end{equation}
This setting leads to an efficient scheme for estimating the temporal sources through stochastic variational inference (Section \ref{ssec:vif}).
\subsubsection{Spatial sources.}
\label{sssec:spatial_sources}
According to the model introduced in Section \ref{ssec:data_modelling}, each observation $\vect{Y}_{p:}$ is obtained as the linear combination at a specific time-point between the temporal and spatial sources. In order to deal with the multi-scale nature of the imaging signal, we propose to represent the spatial sources at multiple resolutions. To this end, we encode the spatial sources in a matrix $\vect{A}$ whose rows $\vect{A}_{n:}$ represent a specific source at a given scale. The scale is prescribed by a convolution operator $\vect{\Sigma}^{n}$, which is a applied to a map $\vect{B}_{n:}$ that we wish to infer. This problem can be specified by defining $\vect{A}_{n:} = \vect{B}_{n:}\vect{\Sigma}^{n}$, where $\vect{\Sigma}^{n}$ is an $F \times F$ Gaussian kernel matrix imposing a specific spatial resolution. The length-scale parameter $\lambda_{n}$ of the Gaussian kernel is fixed for each source, to force the model to pick details at that specific scale. Due to the high-dimension of the data we are modeling, performing stochastic variational inference in this setting raises scalability issues. For instance, if we assume a  Gaussian distribution $\mathcal{N}(\mu_{\vect{B}_{n:}}, diag(\vect{\Lambda}))$ for $\vect{B}_{n:}$, the distribution of the spatial signal would be $p(\vect{A}_{n:}) \sim \mathcal{N}(\vect{\mu}_{\vect{B}_{n:}}\vect{\Sigma}^{n},\vect{\Sigma}^{n}diag(\vect{\Lambda})(\vect{\Sigma}^{n})^{T})$. As a result, sampling from $p(\vect{A}_{n:})$ is not computationally tractable due to the size of the covariance matrix, which prevents the use of standard inference schemes on $\vect{B}_{n:}$. This can be overcome thanks to the separability of the Gaussian convolution kernel \cite{ref_marquand,ref_kronecker}, according to which the 3D convolution matrix $\vect{\Sigma}^{n}$ can be decomposed into the Kronecker product of 1D matrices, $\vect{\Sigma}^{n}=\vect{\Sigma}_{x}^{n} \otimes \vect{\Sigma}_{y}^{n} \otimes \vect{\Sigma}_{z}^{n}$. This decomposition allows to efficiently perform standard operations such as matrix inversion, or matrix-vector multiplication \cite{ref_saatci}. Thanks to this choice, we recover tractability for the inference of $\vect{B}_{n:}$ through sampling, as required by stochastic inference methods \cite{ref_kingma}. 

\subsection{Sparsity}
\label{ssec:sparsity}
In order to detect specific brain areas involved in neurodegeneration, we propose to introduce a sparsity constraint on the maps (or codes) $\vect{B}_{n:}$. Consistently with our variational inference scheme, we induce sparsity via \textit{Variational Dropout} as proposed in \cite{ref_kingma_vd}. This approach leverages on an improper log-scale uniform prior $p(|\vect{B}_{n:}|) \propto \prod_{f} \  1/|\vect{B}_{n,f}|$, along with an approximate posterior distribution:
\begin{equation}
q_{1}(\vect{B}) = \displaystyle \prod_{n=1}^{N} \mathcal{N}(\vect{M}_{n:}, diag(\alpha_{n,1}\vect{M}_{n,1}^{2}... \alpha_{n,F}\vect{M}_{n,F}^{2})).
\end{equation} 
\newline
In this formulation, the dropout parameter $\alpha_{n,f}$ is related to the individual dropout probability $p_{n,f}$ of each weight by $\alpha_{n,f} = p_{n,f}(1-p_{n,f})^{-1}$. When the parameter $\alpha_{n,f}$ exceeds a fixed threshold, the dropout probability $p_{n,f}$ is considered high enough to ignore the corresponding weight $\vect{M}_{n,f}$ by setting it to zero. However, this framework raises stability issues affecting the inference of the dropout parameters due to large-variance gradients, thus limiting $p_{n,f}$ to values smaller than $0.5$. To tackle this problem, we leverage on the extension of \textit{Variational Dropout} proposed in \cite{ref_molchanov}. In this setting, the variance parameter is encoded in a new independent variable $\vect{P}_{n,f} = \alpha_{n,f}\vect{M}^{2}_{n,f}$, while the posterior distribution is optimized with respect to ($\vect{M}, \vect{P}$). Therefore, in order to minimize the cost function for large variance $\vect{P}_{n,f} \to\infty$ ($\alpha_{n,f} \to\infty$ i.e $p_{n,f} \to 1$), the value of the weight's magnitude must be controlled by setting to zero the corresponding parameter $\vect{M}_{n,f}$. As a result, by dropping out weights in the code, we sparsify the estimated spatial maps, thus better isolating relevant spatial sub-structures. Spatial correlations in the images are obtained thanks to the convolution operation detailed in Section \ref{sssec:spatial_sources}. 

\subsection{Variational inference}
\label{ssec:vif}
We detailed in the previous sections the choices of priors and constraints that we apply to the spatio-temporal processes in order to plausibly model the data. To illustrate the overall formulation of the method, we provide in Figure \ref{fig:graphical_model} the graphical model over the M modalities in the case of imaging data. Naturally, this graph simplifies when we deal with scalar data as we don't need to account for any spatial dependence.
\begin{figure}[h!]
\centering
\includegraphics[width=.75\textwidth]{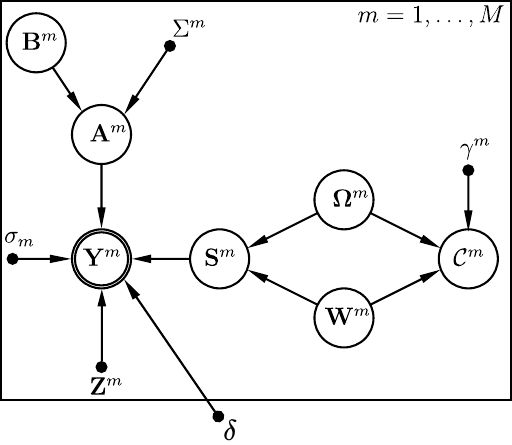}
\caption{Graphical model for imaging data, $\vect{Y}= \{\vect{Y}^{m}\}$.}
\label{fig:graphical_model}
\end{figure}
To infer the time-shift parameter $\vect{\delta}$, the sets of parameters $\theta_m$, $\theta_c$, and $\psi_m$, as well as $\vect{Z}$, $\sigma$ and $\nu$, we need to jointly optimize the data evidence according to priors and constraints:
\begin{align}
\begin{split}
\label{eq:marginal} 
\displaystyle 
\log(p(\vect{Y}, \vect{V}, \mathcal{C}|\vect{Z}, \vect{\delta}, \sigma, \nu, \gamma)) & = \sum_{m} \log(p(\vect{Y}^{m}, \mathcal{C}^{m}|\vect{Z}^{m}, \vect{\delta}, \sigma_{m}, \gamma_{m})) + \sum_{c} \log(p(\vect{V}_{:c}, \mathcal{C}^{c}|\vect{\delta},\nu_{c}, \gamma_{c})). 
\end{split}
\end{align}
We tackle the optimization of Equation \eqref{eq:marginal} via stochastic variational inference. Following \cite{ref_gp} and \cite{ref_monotonicity} we introduce approximations, $q_2(\vect{\Omega}^{m})$ and $q_{3}(\vect{W}^{m})$ in addition to $q_1(\vect{B}^{m})$ in order to derive a lower bound $\vect{\mathcal{L}}_{m}$ for each modality. We recall that the temporal trajectories $\vect{S}^{m}$ and $\vect{U}$ are treated similarly as described in Section \ref{sssec:temporal_sources}. We also note that the choice of distributions $q_1, q_2$ and $q_3$ is the same across modalities, while their parameters will be inferred independently. This leads to:
\begin{align}
\begin{split}
\label{eq:elbo_m}
\log(p(\vect{Y}^{m},\mathcal{C}^{m}|\vect{Z}^{m}, \vect{\delta}, \sigma_{m}, \gamma_{m})) \geqslant & \E_{q_1, q_2, q_3}[\log(p(\vect{Y}^{m}|\vect{B}^{m},\vect{\Omega}^{m}, \vect{W}^{m}, \vect{Z}^{m}, \vect{\delta}, \sigma_{m}))] \\
& + \E_{q_2,q_3}[\log(p(\mathcal{C}^{m}|\vect{\Omega}^{m},\vect{W}^{m}, \vect{\delta}, \gamma_{m}))] \\ 
& - \mathcal{D}[q_{1}(\vect{B}^{m})||p(\vect{B}^{m})] - \mathcal{D}[q_{2}(\vect{\Omega}^{m})||p(\vect{\Omega}^{m})] - \mathcal{D}[q_{3}(\vect{W}^{m})||p(\vect{W}^{m})], \\
\log(p(\vect{V}_{c:},\mathcal{C}^{c}|\vect{\delta}, \nu_{c}, \gamma_{c})) \geqslant & \E_{q_2,q_3}[\log(p(\vect{V}_{c:}|\vect{\Omega}^{c}, \vect{W}^{c}, \vect{\delta}, \sigma_{c}))] \\
& + \E_{q_2,q_3}[\log(p(\mathcal{C}^{c}|\vect{\Omega}^{c},\vect{W}^{c}, \vect{\delta}, \gamma_{c}))] \\ 
& - \mathcal{D}[q_{2}(\vect{\Omega}^{c})||p(\vect{\Omega}^{c})] - \mathcal{D}[q_{3}(\vect{W}^{c})||p(\vect{W}^{c})]
\end{split}
\end{align}
Where $\mathcal{D}$ refers to the Kullback-Leibler (KL) divergence. Combining the lower bounds of the different modalities we obtain:
\begin{align}
\begin{split}
\label{eq:total_elbo} 
\displaystyle 
\log(p(\vect{Y}, \vect{V}, \mathcal{C}|\vect{Z}, \vect{\delta}, \sigma, \nu, \gamma)) & \geqslant \sum_{m} \vect{\mathcal{L}}_{m} + \sum_{c} \vect{\mathcal{L}}_{c}.
\end{split}
\end{align}
A detailed derivation of the lower bound is given in Appendix A.
\newline
The approximated distributions $q_{2}(\vect{\Omega}^{m})$ and $q_{3}(\vect{W}^{m})$ are factorized across GPs such that:
\begin{align}
\begin{split}
 q_{2}({\vect{\Omega}}^{m}) & = \prod_{n=1}^{N_{m}}q_{2}({\vect{\omega}^{n}})^{m} = \prod_{n=1}^{N_{m}}\prod_{j=1}^{N_{rf}} \mathcal{N}(\vect{R}_{n,j}, \vect{Q}_{n,j}^{2})^{m}, \\
 q_{3}({\vect{W}}^{m}) & = \prod_{n=1}^{N_{m}}q_{3}({\vect{w}^{n}})^{m} = \prod_{n=1}^{N_{m}}\prod_{j=1}^{N_{rf}} \mathcal{N}(\vect{T}_{n,j}, \vect{V}_{n,j}^{2})^{m},
\end{split}
\end{align}
where $N_{rf}$ is the number of random features used for the projection in the spectral domain. Using Gaussian priors and approximations we introduced above, we can obtain a closed-form formula for the KL divergence. Moreover, the choice of prior and approximate posterior distribution for the maps of $\vect{B}^{m}$ leads to an approximation for the divergence $\mathcal{D}[q_{1}(\vect{B}^{m})||p(\vect{B}^{m})]$ detailed in \cite{ref_molchanov}. This allows to analytically compute all the KL terms in our cost function. Formulas for the KL divergences are detailed in Appendix B. 
\newline
\newline
Finally, we optimize the individual time-shifts $\vect{\delta} = \{\delta_p\}_{p=0}^{P}$, $\vect{Z}$, $\sigma = \{\sigma_m\}_{m=1}^{M}$, $\nu = \{\nu_c\}_{c=1}^{C}$ as well as the overall sets of spatio-temporal parameters $\vect{\theta} = \{\theta_m\}_{m=1}^{M} \cup \{\theta_c\}_{c=1}^{C}$ and $\vect{\psi}  = \{\psi_m\}_{m=1}^{M}$.
\begin{align}
\begin{split}
    \vect{\theta} & = \{\vect{R}_{n:}^{m}, \vect{Q}_{n:}^{m}, \vect{T}_{n:}^{m}, \vect{V}_{n:}^{m}, l_{n}, n \in [1, N_{m}]\}_{m=1}^{M} \cup \{\vect{R}_{c:}, \vect{Q}_{c:}, \vect{T}_{c:}, \vect{V}_{c:}^{c}, l_{c}\}_{c=1}^{C}, \\
    \vect{\psi}  & = \{\vect{M}_{n:}^{m}, \vect{P}_{n:}^{m}, n \in [1, N_{m}]\}_{m=1}^{M}.
\end{split}
\end{align}

Following \cite{ref_kingma} and using the reparameterization trick, we can efficiently sample from the approximated distributions $q_{1}, q_{2}$ and $q_{3}$ to compute the two expectation terms from \eqref{eq:elbo_m} for each modality. We chose to alternate the optimization between the spatio-temporal parameters and the time-shift. We set $\gamma_m$ to the minimum value that gives monotonic sources. This was done through multiple tests on data batches with different numbers of imaging features $F_m$ and sources $N_m$. We empirically found that monotonicity was enforced when the magnitude of $\gamma_m$ was in the order of $F_m \times N_m$. The threshold for the dropout probability above which we set a weight $\vect{B}^{m}_{n,f}$ to zero was fixed at $95\%$ (i.e $\alpha = 19$), while  the $\sigma_m$ and $\nu_m$ were optimized during training along with the spatio-temporal parameters. The model is implemented and trained using the Pytorch library \cite{ref_pytorch}. The complete experimental setting is detailed in Appendix C. We also provide a pseudo-code detailing the optimization procedure in Appendix D. In the following sections we will refer to our method as Monotonic Gaussian Process Analysis (MGPA).

\section{Experiments and Results}
\label{sec:results}
\documentclass[main.tex]{subfiles}
In this section we first benchmark MGPA on synthetic data to demonstrate its reconstruction and separation properties while comparing it to standard sources separation methods. We finally apply our model on a large set of medical data from a publicly available clinical study, demonstrating the ability of our method to retrieve spatio-temporal processes relevant to AD, along with a time-scale describing the course of the disease. 

\subsection{Synthetic tests on spatio-temporal trajectory separation}
\label{ssec:synthetic_tests}
For the synthetic tests we considered the case where the data is associated to a single imaging modality only. We tested MGPA on synthetic data generated as a linear combination of temporal functions and 3D activation maps at prescribed resolutions. The goal was to assess the method's ability to identify the spatio-temporal sources underlying the data. We benchmarked our method with respect to ICA, Non-Negative Matrix Factorization (NMF), and Principal Component Analysis (PCA), which were applied from the standard implementation provided in the Scikit-Learn library \cite{ref_sklearn}.
\newline
\newline
The benchmark was specified by defining a $10$-folds validation setting, generating the data at each fold as a linear combination of temporal sources $ \widetilde{\vect{S}}(\vect{t}) = [\widetilde{\vect{S}}_{:0}(\vect{t}), \widetilde{\vect{S}}_{:1}(\vect{t})]$, and spatial maps $ \widetilde{\vect{A}} =[\widetilde{\vect{A}}_{0:}, \widetilde{\vect{A}}_{1:}]$. The data was defined as $\vect{Y}_{p:} = \widetilde{\vect{S}}_{p:}(\vect{t}_{p})\widetilde{\vect{A}} + \vect{\mathcal{E}}_{p:}$ over $50$ time points $\vect{t}_{p}$, where $\vect{t}_{p}$ was uniformly distributed in the range $[0, 0.7]$, and $\vect{\mathcal{E}}_{p:} \sim \mathcal{N}(\vect{0}, \sigma^{2}\vect{I})$. The temporal sources were specified as sigmoid functions $\widetilde{\vect{S}}_{p,i}(\vect{t}_{p}) = 1/(1 + \exp(-\vect{t}_{p} + \alpha_{i}))$, while the spatial structures had dimensions $(30 \times 30 \times 30)$ such that $\widetilde{\vect{A}}_{i:} = \widetilde{\vect{B}}_{i:}\widetilde{\vect{\Sigma}}^{i}$. The $\widetilde{\vect{\Sigma}}^{i}$ were chosen as Gaussian convolution matrices with respective length-scale of $\lambda = 2$ mm and $\lambda = 1$ mm. The $\widetilde{\vect{B}}_{i:}$ were randomly sampled sparse 3D maps. 
\newline
\newline
\textbf{Variable selection.} We applied our method by specifying an over-complete set of six sources with respective spatial length-scale of $\lambda = \{2, 2, 1, 1, 0.5, 0.5 \ mm \}$. Figure \ref{fig:sparse_synthetic} shows an example of the sparse maps obtained for a specific fold.  The model prunes the signal for most of the maps, while retaining two sparse maps, $\vect{B_{0:}}$ and $\vect{B_{4:}}$, whose length-scale are $\lambda = 2$ mm and $\lambda = 1$ mm, thus correctly estimating the right number of sources and their spatial resolution. As it can be qualitatively observed in Figure \ref{fig:sparse_synthetic}, we notice that the estimated sparse code convolved with a Gaussian kernel matrix with $\lambda = 1$ mm is closer to its ground truth than the one convolved with a length-scale $\lambda = 2$ mm. According to our tests, sparse codes associated to high resolution details (low $\lambda$) are indeed more identifiable. On the contrary, the identifiability of images obtained via a convolution operator with larger kernels (large $\lambda$) is lower, since these maps can be equivalently obtained through the convolution of different sparse codes.
\begin{figure}[h!]
\centering
\includegraphics[width=1.\textwidth]{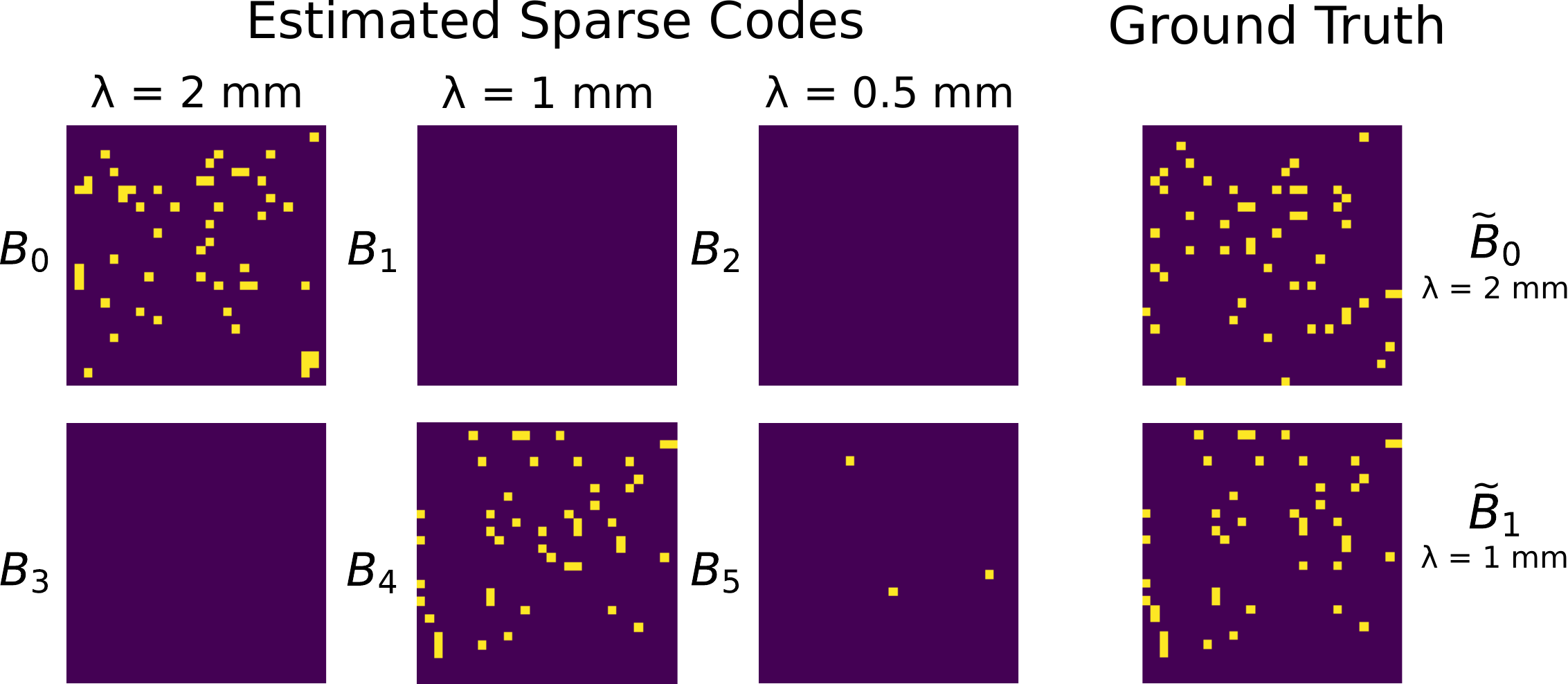}
\caption{Slices extracted from the six sparse codes and the ground truth. Blue: Rejected points. Yellow: Retained points.}
\label{fig:sparse_synthetic}
\end{figure}
\newline
\newline
\textbf{Sources separation.} We observe in Table \ref{mse_table} that the lowest Mean-Squared Error (MSE) for the temporal sources reconstruction is obtained by MGPA, closely followed by ICA. Similarly, our model and ICA show the highest Structural Similarity (SSIM) score \cite{ref_ssim}, which quantifies the image reconstruction accuracy with respect to the ground truth maps, while accounting for the inter-dependencies between neighbouring pixels. An example of image reconstruction from a sample fold is illustrated in Figure \ref{fig:ica_comparison}. In this standard benchmark, we note that MGPA leads to comparable results with respect to the state of the art. In the following section, we compare the models in the more challenging setting in which the time-line has to be estimated as well. 
\begin{table}[h!]
\caption{MSE and SSIM between respectively the ground truth temporal and spatial sources with respect to the ones estimated by the different methods.}
\label{mse_table}
\vskip 0.15in
\begin{center}
\begin{small}
\begin{sc}
\begingroup
\setlength{\tabcolsep}{10pt}
\begin{tabular}{ccc}
\toprule
 & Temporal (MSE) & Spatial (SSIM) \\
\midrule
MGPA & $\mathbf{(8\pm 4).10^{-5}}$ & $\mathbf{98\% \pm 1}$ \\
ICA & $ (6\pm 3).10^{-4}$ & $97 \% \pm 2$ \\
NMF & $(3\pm2).10^{-2}$ & $40 \% \pm 17$ \\
PCA & $0.44 \pm 10^{-3}$ & $15 \% \pm 1$  \\
\bottomrule
\end{tabular}
\endgroup
\end{sc}
\end{small}
\end{center}
\vskip -0.1in
\end{table}

\begin{figure}[h!]
\centering
\includegraphics[width=.9\textwidth]{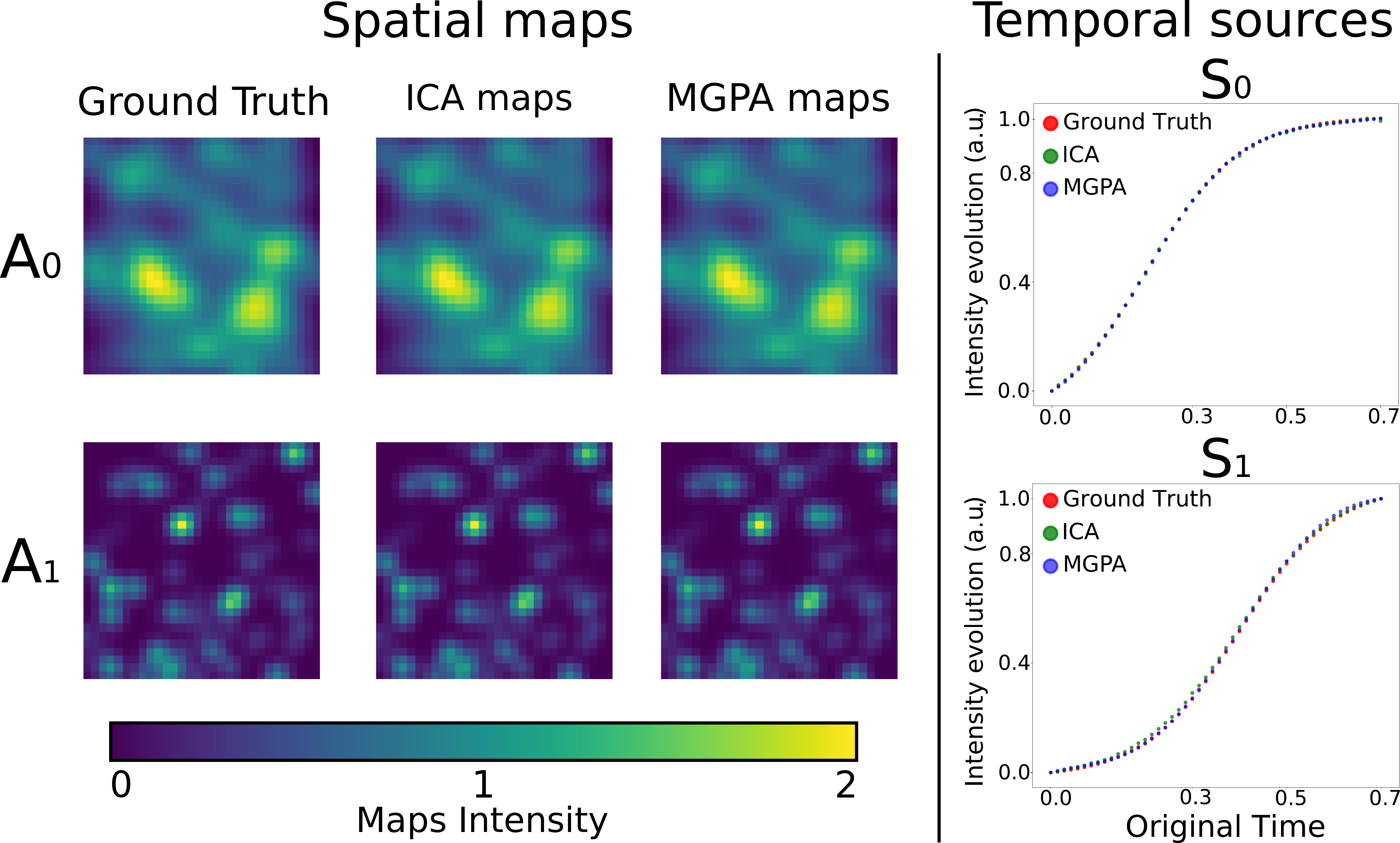}
\caption{Spatio-temporal reconstruction when inference on the time-line is not required. Spatial maps: Sample slice from ground truth images ($A_0$ $\lambda = 2$ mm, $A_1$ $\lambda = 1$ mm), the maps estimated by ICA, and the ones estimated by MGPA. Temporal sources: Ground truth temporal sources (red) along with sources estimated by ICA (green) and MGPA (blue).}
\label{fig:ica_comparison}
\end{figure}

\subsection{Synthetic tests on trajectory separation and time-reparameterization}
\label{ssec:synthetic_shift}
In this test, we modify the experimental benchmark by introducing a further element of variability associated to the time-axis. The temporal and spatial sources were modelled following the same procedure as in Section \ref{ssec:synthetic_tests}, however the observations were mixed along the temporal axis. To do so we generated longitudinal data as $\vect{Y}_{p,j,:} = \widetilde{\vect{S}}_{p:}(\vect{t})\widetilde{\vect{A}} + \vect{\mathcal{E}}_{j:}$, by sampling between $1$ and $10$ images per time-point and randomly re-arranging them along the time-axis (cf. time-shift $t_p$ of each observation at initialization in Figures \ref{fig:sources_separation1} and \ref{fig:sources_separation2}, panel ``Time-Shift''). The goal was to assess the sources separation performances of MGPA when the time-line is unknown. The experiment was run on 10 folds and Figures \ref{fig:sources_separation1} and \ref{fig:sources_separation2} illustrate the sources estimation for two different folds. We present these two figures to demonstrate how the time-shift inference affects the temporal sources reconstruction. Since the model is agnostic of a time-scale, we note that the time-shift may have a different range than the original time-axis. However, its relative ordering should be consistent with the original time points. We fitted a linear regression model over the 10 folds between the original time and the estimated time-shift parameter, and obtained an average R$^{2}$ coefficient of $0.98$ with a standard deviation of $0.005$ (cf. Table \ref{time_shift_table}). 
\begin{table}[h!]
\caption{MSE and SSIM between respectively the ground truth temporal and spatial sources with respect to the ones estimated by MGPA. R$^{2}$ coefficient of the linear regression between the original time-line and the estimated time-shift.}
\label{time_shift_table}
\vskip 0.15in
\begin{center}
\begin{small}
\begin{sc}
\begingroup
\setlength{\tabcolsep}{10pt}
\begin{tabular}{cccc}
\toprule
 & Temporal (MSE) & Spatial (SSIM) & R$^{2}$  \\
\midrule
MGPA & $\mathbf{(2\pm 0.8).10^{-2}}$ & $\mathbf{95\% \pm 4}$ & $\mathbf{0.98 \pm 0.005}$ \\
\bottomrule
\end{tabular}
\endgroup
\end{sc}
\end{small}
\end{center}
\vskip -0.1in
\end{table}
This is illustrated for two different folds in the Time-Shift panel of Figures \ref{fig:sources_separation1} and \ref{fig:sources_separation2}, where we observe a strong linear correlation with the original time-line, meaning that the algorithm correctly re-ordered the data with respect to the original time-axis. However, we notice in Table \ref{time_shift_table} that the MSE of the temporal sources significantly increased, due to the additional difficulty brought by the time-shift estimation. Indeed, in order to reconstruct the temporal signal we need to perfectly re-align hundreds of observations. This is the case in Figure \ref{fig:sources_separation1} (optimal reconstruction result), where the time-shift is highly correlated with the original time-line, allowing to distinguish every single observation and reconstruct the original temporal profiles. Whereas in Figure \ref{fig:sources_separation2} (sub-optimal reconstruction result), the estimated time-shift doesn't exhibit a perfect fit, and generally underestimates the time-reparameterization for the later and earlier time points. This is related to the challenging setting of reconstructing the time-line identified by the original temporal sources. Indeed, we observe that $\vect{S}_{:0}$ reaches a plateau for early time points, while $\vect{S}_{:1}$ is flat for later ones. This behaviour increases the difficulty of differentiating time points with low signal differences. As a result, it impacts the time-shift optimization and adds variability to the time-shift estimation performances, thus deteriorating the reconstruction of the temporal sources over the 10 folds compared to the previous benchmark. The spatial sources estimation remains comparable to the one without time-shift both quantitatively, with an average SSIM of 95$\%$, and qualitatively, as shown in Figures \ref{fig:sources_separation1} and \ref{fig:sources_separation2}. Within this setting, ICA, NMF and PCA poorly perform as they can't reconstruct the time-line. Results obtained using these three methods are provided in Appendix E.  
\begin{figure}[h!]
\centering
\includegraphics[width=.9\textwidth]{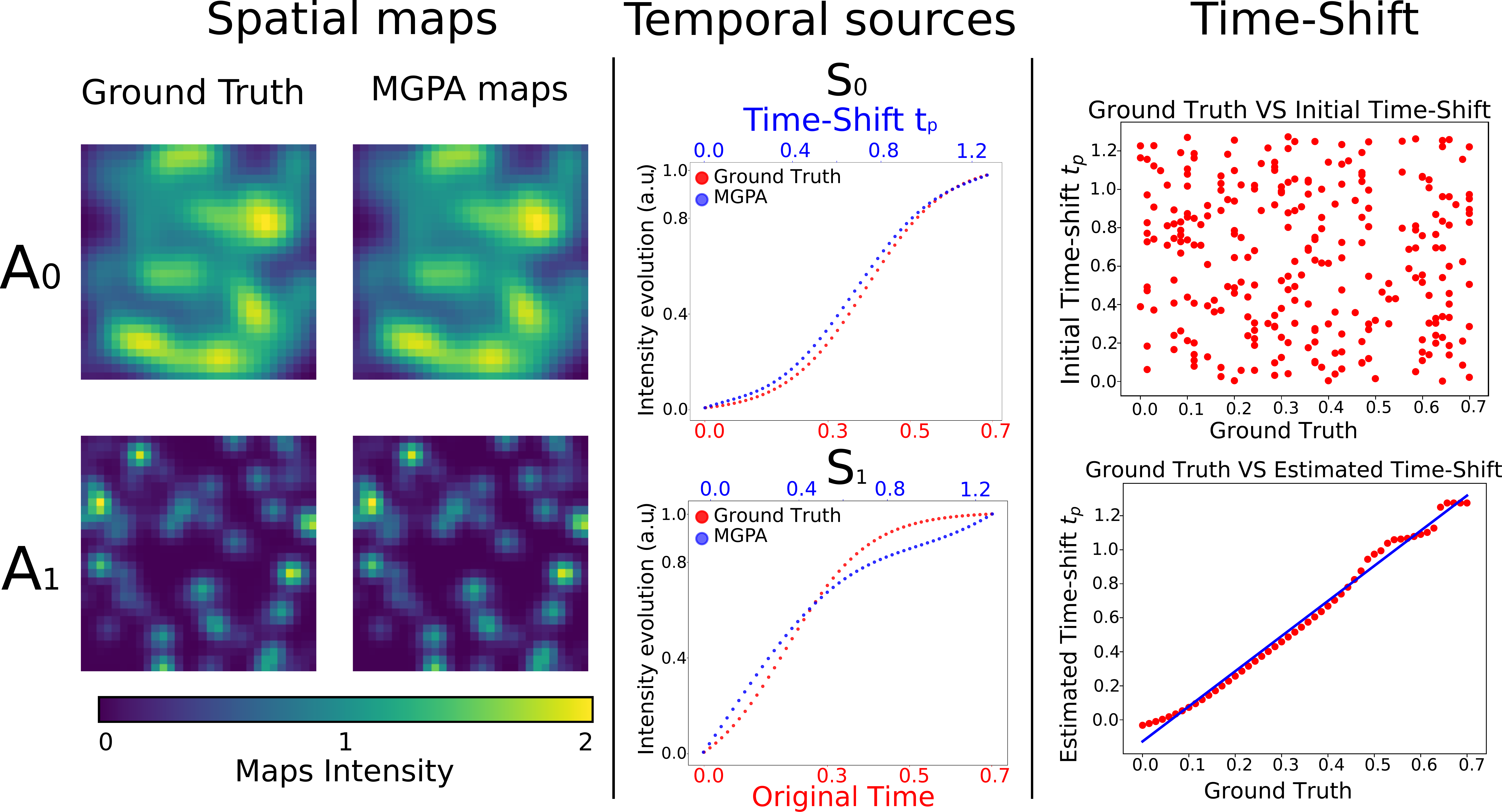}
\caption{Spatio-temporal reconstruction when inference on the time-line is required. Optimal reconstruction result. Spatial maps: Sample slice from ground truth images ($A_0$ $\lambda = 2$ mm, $A_1$ $\lambda = 1$ mm) and estimated spatial sources. Temporal sources: In red the original temporal sources, in blue the estimated temporal sources. Time-Shift: Time-shift $t_p$ of each image at initialization (top), and after estimation (bottom). In blue, linear fit with the ground truth.}
\label{fig:sources_separation1}
\end{figure}
\begin{figure}[h!]
\centering
\includegraphics[width=.9\textwidth]{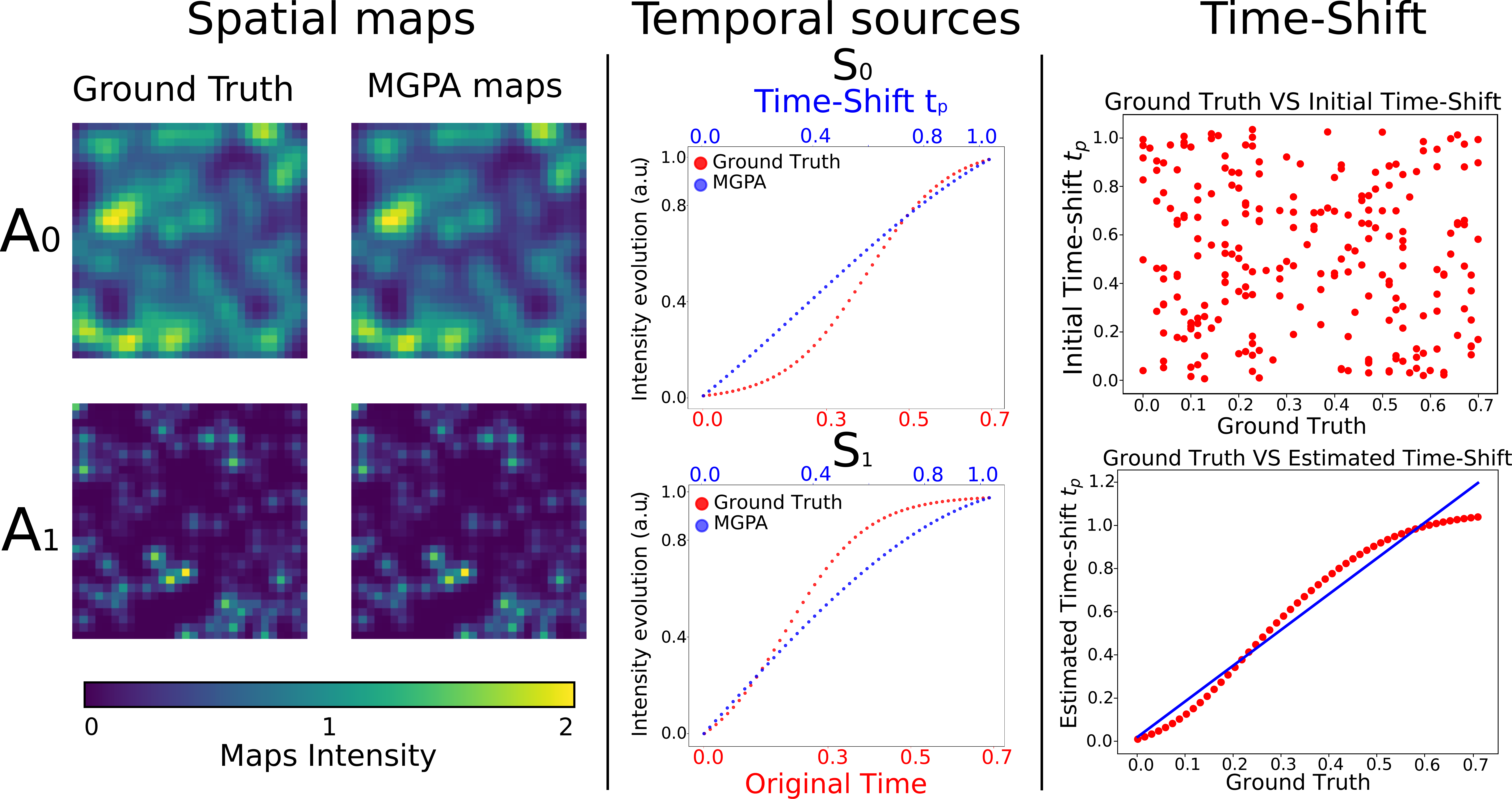}
\caption{Spatio-temporal reconstruction when inference on the time-line is required. Sub-optimal reconstruction result. Spatial maps: Sample slice from ground truth images ($A_0$ $\lambda = 2$ mm, $A_1$ $\lambda = 1$ mm) and estimated spatial sources. Temporal sources: In red the original temporal sources, in blue the estimated temporal sources. Time-Shift: Time-shift $t_p$ of each image at initialization (top), and after estimation (bottom). In blue, linear fit with the ground truth.}
\label{fig:sources_separation2}
\end{figure}
\subsection{Application to spatio-temporal brain progression modeling}
\label{ssec:results_real_data}
\subsubsection{Data processing}
\label{sssec:data_preprocessing}
Data  used in the preparation of this article were obtained from the Alzheimer’s Disease Neuroimaging Initiative (ADNI) database (adni.loni.usc.edu). The ADNI was launched in 2003 as  a public-private partnership, led by  Principal Investigator Michael W. Weiner, MD. For up-to-date information, see www.adni-info.org.
\newline
\newline
We selected a cohort of 544 amyloid positive subjects of the ADNI database composed of 103 controls (NL), 164 Mild Cognitive Impairment (MCI), 114 AD patients, 34 healthy individuals converted to MCI or to AD (NL converter) and 129 MCI converted to AD (MCI converter). The term amyloid positive refers to subjects whose amyloid level in the cerebrospinal fluid (CSF) is below the nominal cutoff of $192$ pg/ml. Conversion to MCI or AD was determined using the last follow-up available information. We provide in Table \ref{baseline_info} socio-demographic and clinical information across the different groups.
\begin{table}[h!]
\caption{Baseline socio-demographic and clinical information for study cohort. Average values and standard deviation in parenthesis. NL: normal individuals, NL converter: normal subjects who converted to MCI or to AD, MCI: mild cognitive impairment, MCI converter: MCI subjects who converted to AD, AD: Alzheimer's patients. ADAS13: Alzheimer's Disease Assessment Scale-cognitive subscale, 13 items. FAQ: Functional Assessment Questionnaire. FDG: (18)F-fluorodeoxyglucose Positron Emission Tomography (PET) imaging. AV45: (18)F-florbetapir Amyloid PET imaging.}
\label{baseline_info}
\vskip 0.15in
\begin{center}
\begin{footnotesize}
\begin{sc}
\begingroup
\setlength{\tabcolsep}{10pt}
\begin{tabular}{lccccc}
\toprule
Group & NL & \pbox{20cm}{\hspace{1.5em} NL \\ converter} & MCI & \pbox{20cm}{\hspace{1.1em} MCI \\ converter} & AD  \\
\midrule
N & 103 & 34 & 164 & 129 & 114  \\
Age & 73 (6) & 78 (5) & 73 (7) & 73 (7) & 74 (8) \\
Education {\scriptsize (yrs)} & 16.3 (3) & 16 (3) & 15.7 (3) & 16 (3) & 15.6 (3) \\
ADAS13 & 9.1 (4.4) & 11.4 (4.3) & 14.6 (5.5) & 20.4 (6.5) & 31.6 (8.5) \\
FAQ & 0.3 (0.7) & 0.2 (0.6) & 1.9 (2.8) & 5.0 (4.6) & 13.5 (6.9) \\
Entorhinal {\scriptsize (cm$^{3}$)} & 3.8 (0.5) & 3.5 (0.5) & 3.6 (0.6) & 3.2 (0.7) & 2.8 (0.6) \\
Hippocampus {\scriptsize (cm$^{3}$)} & 7.4 (0.9) & 6.9 (0.7) & 6.9 (0.9) & 6.4 (0.9) & 5.9 (0.8) \\
Ventricles {\scriptsize (cm$^{3}$)} & 31 (16) & 42 (21) & 39 (23) & 40 (19) & 48 (23) \\
Whole brain {\scriptsize (cm$^{3}$)} & 1033 (104) & 1019 (91) & 1058 (103) & 1037 (102) & 1005 (115) \\
FDG & 1.3 (0.1) & 1.3 (0.1) & 1.2 (0.1) & 1.1 (0.1) & 1.0 (0.1) \\
AV45 & 1.3 (0.2) & 1.3 (0.1) & 1.3 (0.2) & 1.4 (0.2) & 1.5 (0.2) \\
\bottomrule
\end{tabular}
\endgroup
\end{sc}
\end{footnotesize}
\end{center}
\vskip -0.1in
\end{table}
\newline
\newline
MRI, FDG-PET and AV45-PET of each individual were processed in order to obtain respectively, volumes of gray matter density, glucose uptake, and amyloid load in a standard anatomical space.
\newline
\newline
\textbf{MRI processing protocol.} Baseline MRI images were analyzed according to the SPM12 processing pipeline \cite{ref_spm}. Each image was initially segmented into grey, white matter and CSF probabilistic maps. Grey matter images were used for the following analysis, normalized to a group-wise reference space via DARTEL \cite{ref_dartel}, and modulated using the Jacobian determinant of the subject-to-template transformation. The subsequent modeling was carried out on the normalised images at the original spatial resolution.
\newline
\newline
\textbf{PET processing protocol.} Individuals\textquotesingle \ baseline PET images were initially affinely aligned to the corresponding MRI. After scaling the intensities to the cerebellum, the images were normalized to the grey matter template obtained with DARTEL and smoothed with a FWHM parameter of 4.55.
\newline
\newline
Images have dimension $102 \times 130 \times 107$ before vectorization, leading to $1,418,820$ spatial features per patient. These spatial features represent for each voxel their gray matter concentration in the case of MRI images, their glucose metabolism for FDG-PET images, or their amyloid concentration for AV45-PET images. To exploit the ability of our model to automatically adapt to different spatial scales, we chose to keep the MRI images at their native resolution for the analysis, and thus do not perform additional smoohting to equalize to the PET FWHM. In addition to the imaging data of each patient, we also integrate the ADAS13 score assessed by clinicians. High values of this score indicate a decline of cognitive abilities. We consider three matrices $\vect{Y}^{MRI}$, $\vect{Y}^{FDG}$, and $\vect{Y}^{AV45}$ of dimension $(543 \times 1,418,820)$ containing the images of all the subjects, and a matrix $\vect{V}$ of dimension $(543 \times 1)$ containing their ADAS13 score. From now on we will refer to the data as the block diagonal matrix containing the four matrices $\vect{Y}^{MRI}, \vect{Y}^{FDG}$, $\vect{Y}^{AV45}$, and $\vect{V}$ as described in Section \ref{ssec:data_modelling}. We note that the analysis is performed by only considering a single scan per imaging modality and ADAS13 score for each patient. Therefore, the temporal evolution has to be inferred solely through the analysis of relative differences between the brain morphologies, glucose metabolisms, amyloid concentrations and cognitive abilities across individuals. 

\subsubsection{Model specification}
\label{ssec:model_specification}
We aim at showing how MGPA applied on the data extracted from the ADNI cohort is able to temporally re-align patients in order to describe AD progression in a plausible way, while detecting relevant spatio-temporal processes at stake in AD. The model estimates AD progression by relying on MR, FDG-PET, AV45-PET scans and ADAS13 score of each patient. The temporal sources $\vect{S}^{MRI}$ and $\vect{S}^{FDG}$ associated respectively to the loss of gray matter, and to the decrease of glucose uptake, are enforced to be monotonically decreasing. On the contrary, the temporal sources $\vect{S}^{AV45}$ and $\vect{U}_{:ADAS13}$, modeling respectively the evolution of amyloid concentration, and ADAS13 score, are enforced to be monotonically increasing. Since we don't consider any information about the disease stage of each individual before applying our method, all the observations are initialized at the same time reference $\tau = 0$. Therefore, as for the tests in Section \ref{ssec:synthetic_shift}, the time-shift reparameterization describes a relative re-ordering of the subjects not related to a specific time-unit. To decompose the imaging data we apply our model by specifying an over-complete basis of six sources with $\lambda =\{ 8, 8, 4, 4, 2, 2 \ mm \}$, to cover both different scales and the associated variety of temporal evolution. Due to the high-dimension of the data matrix, the computations were parallelized over six GPUs, and the model required eighteen hours to complete the training. Details on the model convergence during training are provided in Appendix F.

\subsubsection{Estimated spatio-temporal brain dynamics}
\label{sssec:brain_dynamics}
In Figure \ref{fig:full_model} we show the spatio-temporal processes retained by the model for each imaging modality. Interestingly, the model adapts to the spatial resolution of MRI and PET images. Indeed, we notice that the model accounts for the high-resolution of MRI images by retaining a source associated to the lowest length-scale ($\lambda = 2$ mm). Concerning PET data, we observe that the induced sparsity discards the highest resolution codes ($\lambda = 2$ mm) for both FDG and AV45, highlighting the ability of the model to adapt to the coarser resolution of the PET signal. 
\newline
\newline
In the case of MRI data, two sources were retained at two different resolutions ($\lambda = 4$ mm  and $\lambda = 2$ mm). Source $\vect{S}^{MRI}_4$ describes gray matter loss encompassing a large extent of the brain with a focus on cortical areas (see $\vect{A}^{MRI}_{4}$). We note that this map also targets subcortical areas such as the hippocampi, which are key regions of AD. Source $\vect{S}^{MRI}_{2}$ ($\lambda = 4$ mm) indicates a mild decrease of gray matter which accelerates in the latest stages of the disease, and targets the temporal poles (see $\vect{A}^{MRI}_2$). It is interesting to notice that this differential pattern of gray matter loss also affects the parahippocampal region, whose atrophy is known to be prominent in AD \cite{ref_parahippocampal}. These results underline the complex evolution of brain atrophy, and the ability of the model to disentangle spatio-temporal processes mapping different regions involved in the pathology \cite{ref_bateman, ref_frisoni}. Concerning the spatio-temporal processes extracted from the FDG-PET data, we see on Figure \ref{fig:full_model} that the model retained two sources at the coarsest resolutions ($\lambda = 8$ mm). Source $\vect{S}^{FDG}_1$ indicates a pattern of hypometabolism that tends to plateau and which involves most of the brain regions, thus describing a global effect of the pathology on the glucose uptake. Source $\vect{S}^{FDG}_0$ describes a linear pattern of hypometabolism targeting areas such as the precuneus and the parietal lobe, which are known to be strongly affected during the evolution of the disease \cite{ref_brown}. Finally, the model extracted two spatio-temporal sources from the AV45-PET data at two different resolutions ($\lambda = 8$ mm  and $\lambda = 4$ mm). We observe that source $\vect{S}^{AV45}_2$ highlights an increase of amyloid deposition mapping a large extent of the brain, such as the parietal and frontal lobes as well as temporal areas, thus concurring with clinical evidence \cite{ref_amyloid}. Similarly to the FDG-PET processes, we have a source $\vect{S}^{AV45}_0$ exhibiting a differential pattern of amyloid deposition targeting mostly frontal, temporal, occipital areas and precuneus.
\newline
\newline
The estimated spatio-temporal processes can be combined to obtain an estimated evolution $\vect{S}^{m}\vect{A}^{m}$ of the brain along the time-shift axis for each modality. In Figure \ref{fig:ratio_evolution}, we show the ratio $|\vect{S}_{p:}^{m}\vect{A}^{m} -\vect{S}_{0:}^{m}\vect{A}^{m}| / \vect{S}_{0:}^{m}\vect{A}^{m}$ between the image predicted at four time-points $t_{p}$ and the image predicted at $t_0$ for the three imaging modalities. This allows us to visualize the trajectory of a brain going from a healthy to a pathological state in terms of atrophy, glucose metabolism and amyloid load according to our model.
\newline
\newline
Finally, we also applied ICA, NMF and PCA on the ADNI data, showing that the associated results are characterized by poor interpretability and high variability. The complete experimental setting and results are detailed in Appendix G.
\begin{figure}[htbp!]
\centering
\includegraphics[height=1.\textheight]{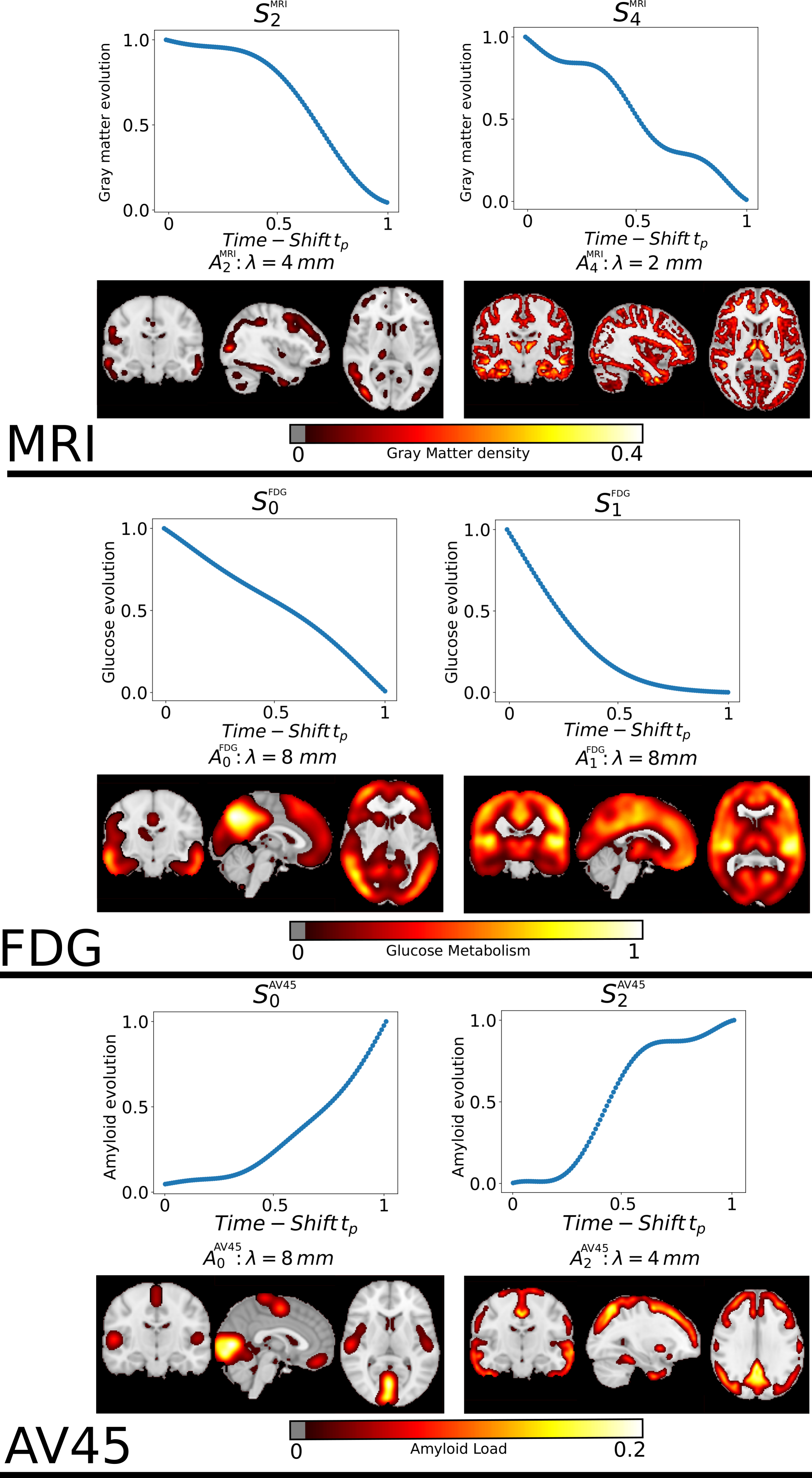}
\caption{Estimated spatio-temporal processes for the three imaging modalities. The time-scale was re-scaled to the arbitrary range [0, 1].}
\label{fig:full_model}
\end{figure}

\begin{figure}[htbp!]
\centering
\includegraphics[width=1.\textwidth]{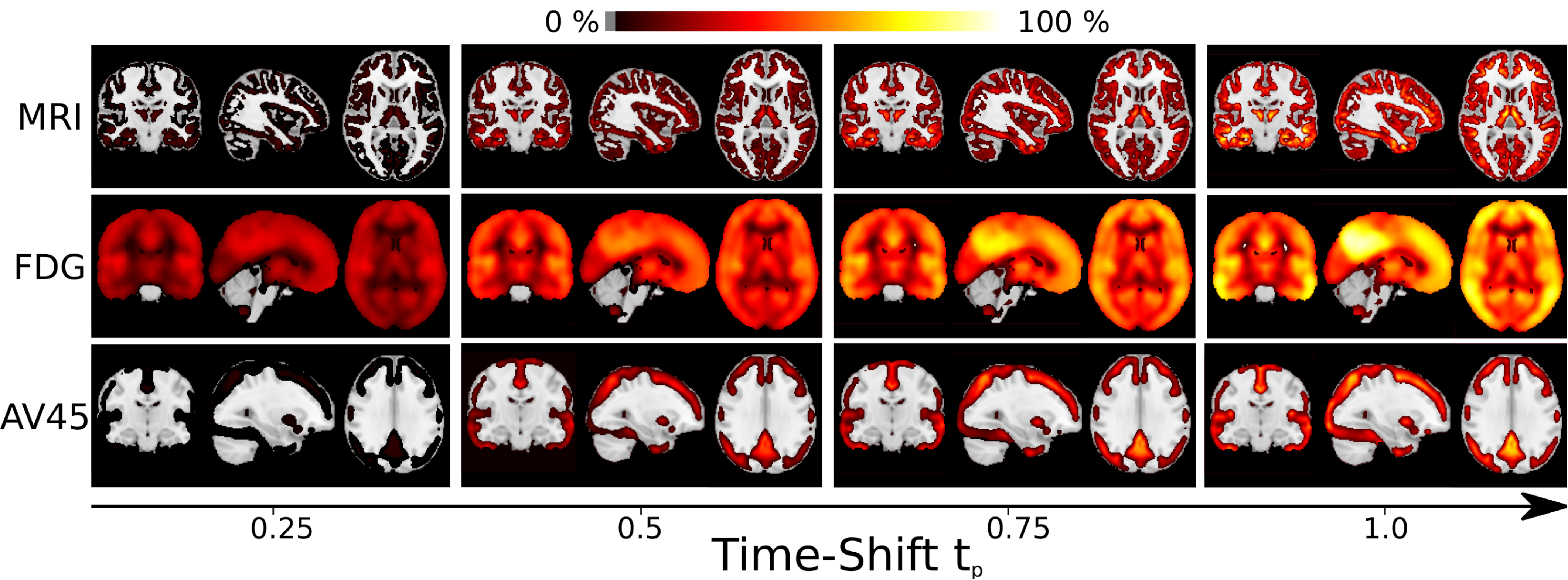}
\caption{Ratio between the model prediction at time $t_p$ and the prediction at $t_0$ for the three imaging modalities. The time-scale was re-scaled to the arbitrary range [0, 1].}
\label{fig:ratio_evolution}
\end{figure}

\subsubsection{Model Consistency}
To verify the plausibility of the fitted model, we compare in Figure \ref{fig:vol_biomarkers} the concentration predicted by the model and the raw concentration measures in different brain areas for the three imaging modalities. We observe a decrease of gray matter and glucose metabolism as we progress along the estimated time-line, allowing to relate large time-shift values to lower gray matter density and glucose uptake. Moreover, we notice the agreement between the predictions made by the model (in blue) and the raw concentration measures (in red). In the case of AV45 data there is only a mild increase of amyloid load according to the model, probably due to the fact that the subjects selected in the cohort are already amyloid positive. As a result, they already show a high baseline amyloid level concentration, close to plateau levels.
\newline
\newline
In Figure \ref{fig:adas13}, we show the estimated GP $\vect{U}_{:ADAS13}$. We observe that the model is able to plausibly describe the evolution of this cognitive score, while demonstrating a larger variability than in the case of imaging modalities. 
\begin{figure}[h!]
\centering
\includegraphics[width=\textwidth]{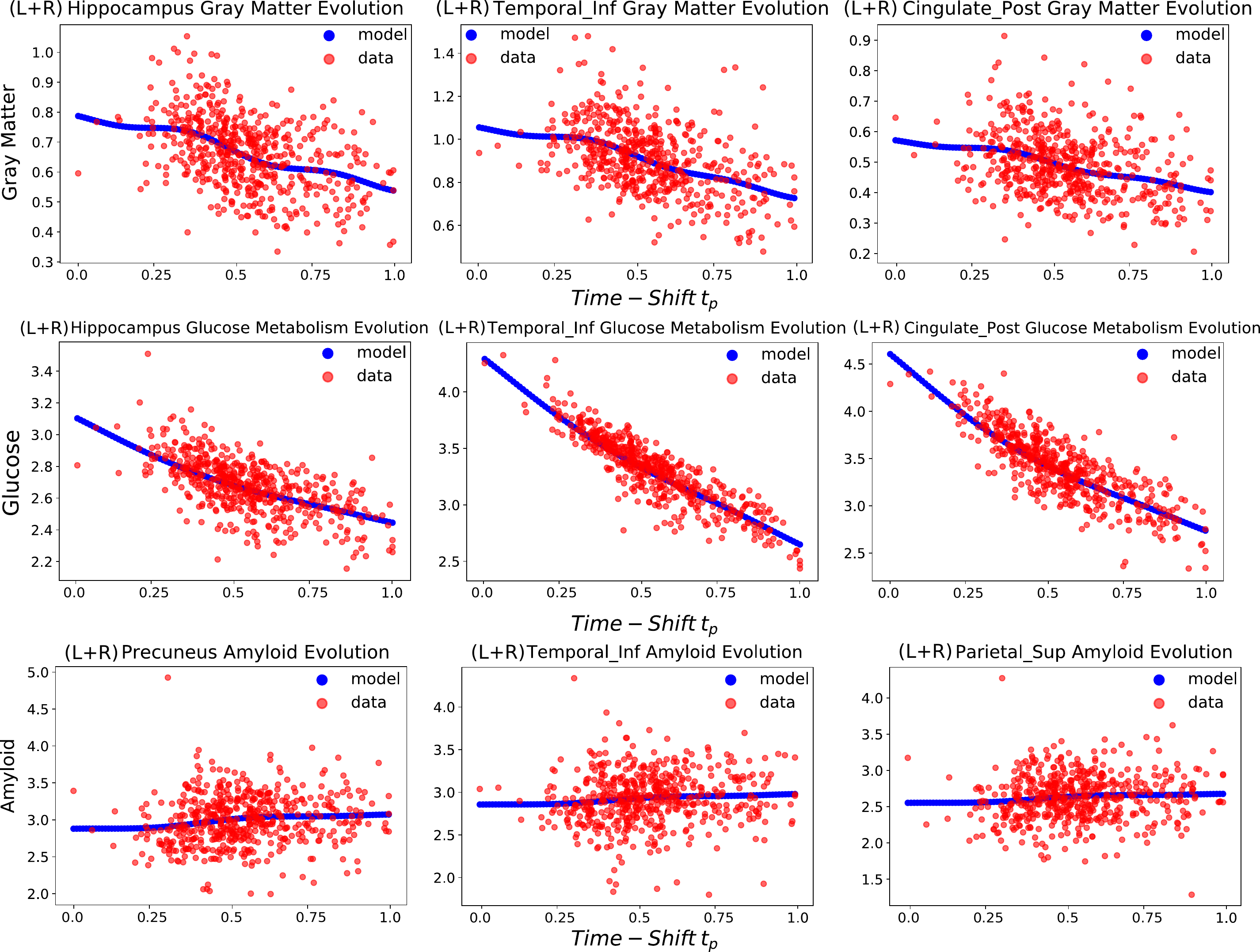}
\caption{Model prediction averaged on specific brain areas (blue line), and observed values (red dots), along the estimated time-line for the three imaging modalities. L and R respectively stand for left and right. The time-scale was re-scaled to the arbitrary range [0, 1].}
\label{fig:vol_biomarkers}
\end{figure}

\begin{figure}[htbp!]
\centering
\includegraphics[width=0.45\textwidth]{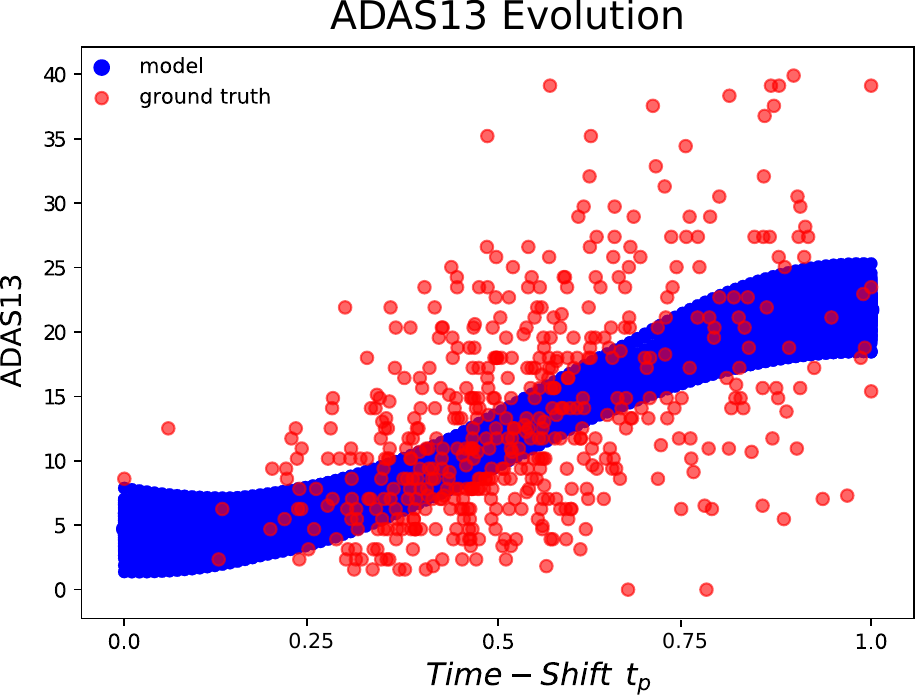}
\caption{Model prediction of the ADAS13 score (blue line), and observed values (red dots) along the estimated time-line. The time-scale was re-scaled to the arbitrary range [0, 1].}
\label{fig:adas13}
\end{figure}

\subsubsection{Plausibility with respect to clinical evidence}
\label{ssec:clinical_validation_adni}
We assessed the clinical relevance of the estimated time-shift by relating it to independent medical information which were not included in the model during training. To this end, we compared the estimated time-shift to ADAS11, MMSE and FAQ scores. High values of ADAS11 and FAQ or low values of MMSE indicate a decline of performances. We show in Figure \ref{fig:clinic} that the estimated time-shift correlates with a decrease of cognitive and functional abilities. In particular, a cubic model slightly better describes the relationship between ADAS11 and the time-shift (according to BIC and AIC), with a significance for the cubic coefficient of $p = 0.04$. Concerning MMSE and FAQ, quadratic and linear models were almost equivalent; the significance of the linear coefficients was $p < 0.01$, while the quadratic coefficient was never significant. Pearson correlation coefficients for ADAS11, FAQ and MMSE were respectively of $0.49$, $0.41$, and $-0.45$, with corresponding p-values $p < 0.01$.
\newline
\newline
The box-plot of Figure \ref{fig:stages_distribution} shows the time-shift distribution across clinical groups. We observe an increase of the estimated time-shift when going from healthy to pathological stages. The high uncertainty associated to the MCI group is due to the broad definition of this clinical category, which includes subjects not necessarily affected by dementia. We note that MCI subjects subsequently converted to AD (MCI converter) exhibit higher time-shift than the clinically stable MCI group, highlighting the ability of the model to differentiate between conversion status. A similar distinction can be noticed between NL and NL converter groups. We found significant differences between median time-shift for NL-NL converter, MCI-MCI converter and MCI converter-AD (comparisons $p < 0.01$, Figure \ref{fig:stages_distribution}). It is also important to recall that this result is obtained from the analysis of a single scan per imaging modality and ADAS13 score for each patient.
\begin{figure}[htbp!]
\centering
\includegraphics[width=0.95\textwidth]{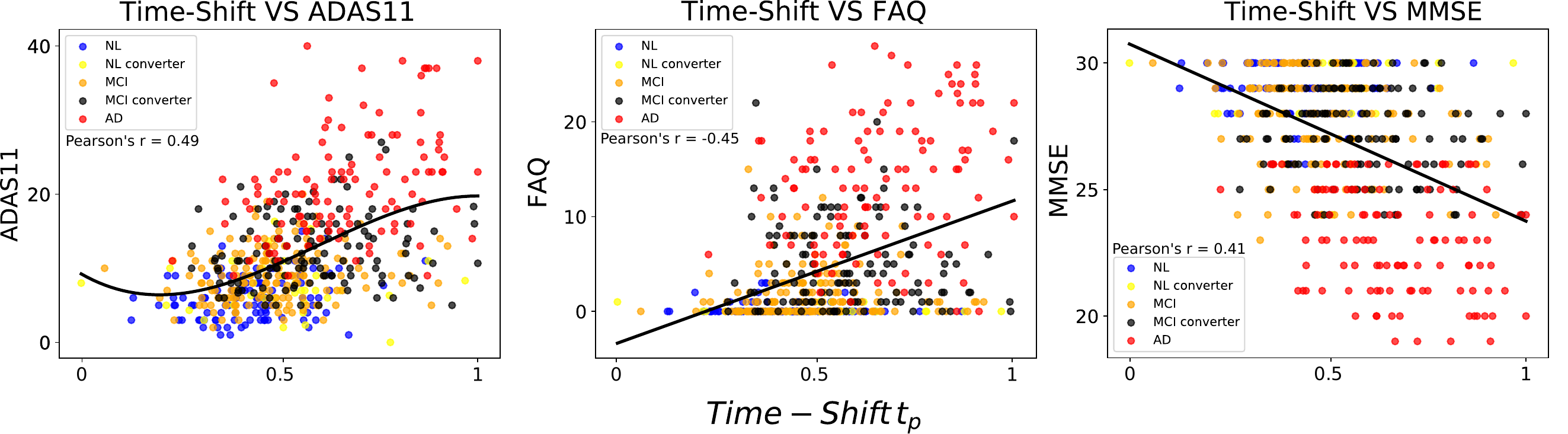}
\caption{Evolution of the ADAS11 (left), FAQ (middle) and MMSE (right) along the estimated time-line. The time-scale was re-scaled to the arbitrary range [0, 1].}
\label{fig:clinic}
\end{figure}

\begin{figure}[htbp!]
\centering
\includegraphics[width=0.5\textwidth]{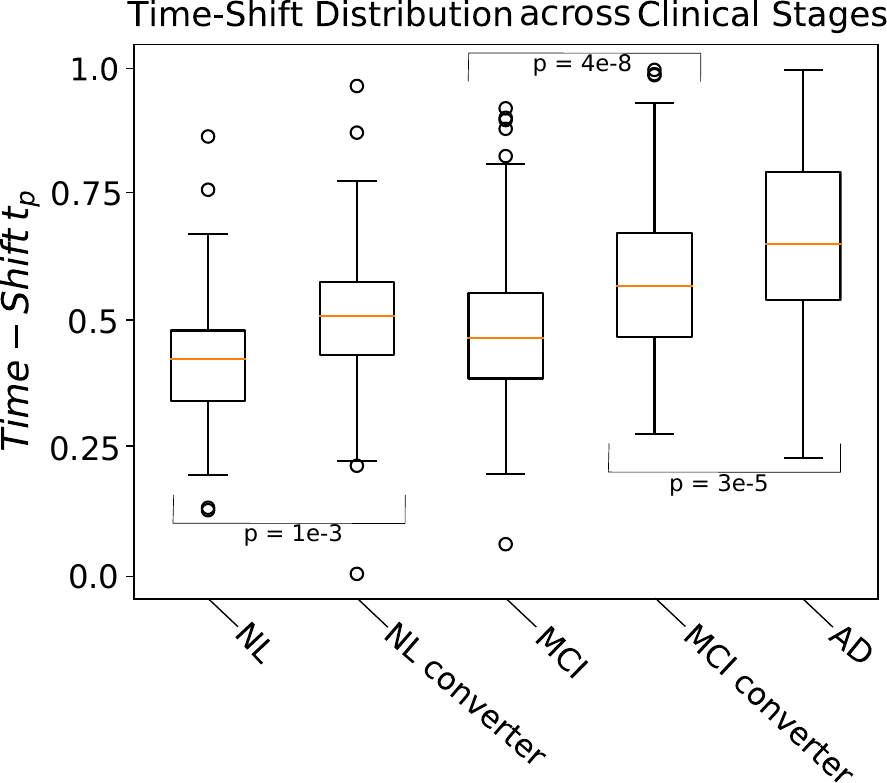}
\caption{Distribution of the time-shift values over the different clinical stages. The time-scale was re-scaled to the arbitrary range [0, 1].}
\label{fig:stages_distribution}
\end{figure}

\section{Discussion}
\label{sec:discussion}
\documentclass[main.tex]{subfiles}
We presented a generative approach to spatio-temporal disease progression modeling based on matrix factorization across temporal and spatial sources. The proposed application on a large set of medical images shows the ability of the model to disentangle relevant spatio-temporal processes at stake in AD, along with an estimated time-scale related to the disease evolution. 
\newline
\newline
The model was compared to standard methods such as ICA, NMF and PCA since they perform blind source separation similarly to our method. This allowed us to demonstrate the advantages of building more complex approaches such as MGPA for the problem we tackle in this work. Concerning the comparison with the state of the art in disease progression modelling, to the best of our knowledge the two closest approaches are \cite{ref_marinescu} and \cite{ref_koval}. However, these two methods are specifically designed for modelling data defined on brain surfaces. On the contrary, our method aims at progression modeling using full 3D volumetric information. The data dimension we tackle is thus an order of magnitude greater than the one of \cite{ref_marinescu} and \cite{ref_koval}, preventing these methods to scale to the spatial geometry of our data.
\newline
\newline
There are several avenues of improvement for the proposed approach. We found that the optimization is highly sensitive to the initialization of the spatial sources. This is typical of such complex non-convex problems, and requires further investigations to better control the algorithm convergence. More generally, the problem of source separation tackled in this work is intrinsically ill-posed, as the given data can be explained by several solutions. This was illustrated for example in our tests on synthetic data (Section \ref{ssec:synthetic_shift}), where the identification of the sources was more challenging in the case of coarse resolution codes and of flat temporal sources. We note however that this issue is general, and intrinsic to the problem of disease progression modeling.
\newline
\newline
Indeed, identifiability ultimately remains a critical issue when training the model. Concerning the spatio-temporal parameters, their number is extremely high due to the fact that we scale our method to 3D volumetric images. Estimating a single spatial source from a single modality requires to estimate the mean and variance of its sparse code, i.e $1,418,820 \times 2 = 2,837,640$ parameters. In practice, hypotheses are explicitly introduced to reduce the number of effective parameters. For instance, the convolution of the spatial maps using Gaussian kernels allows to enforce smoothness, and thus reduces the number of effective degrees of freedom via spatial correlation across the related parameters. This is equivalent to the regularization applied to image registration problems, in which the number of parameters is of the same order of magnitude than in our setting. Moreover, our sparsity constraint allows to sensibly reduce the number of parameters at test time. Indeed, after training, the sparse codes of the MRI sources have $2,213,359$ non-zero elements instead of $17,025,840$, which amounts in $87\%$ reduction in the number of parameters. In the case of the FDG-PET and AV45-PET sparse codes, the number of non-zero elements at test time is respectively of  $9,023,695$ and $1,362,067$, which is equivalent to a reduction in the number of parameters of $53\%$ and $92\%$. Nonetheless, this high number of parameters still remains a factor of potential convergence issues during the parameters estimation procedure. We present graphs in Appendix F showing the evolution of the different terms composing the cost function during training. These figures show convergence profiles typical of those obtained with stochastic variational inference schemes, such as with Variational Autoencoders or Bayesian Neural Networks. Moreover, the stability of the solution has been ensured through multiple runs of the model. Finally, as mentioned in Section \ref{ssec:sparsity}, the \textit{Variational Dropout} framework leads to stability issues affecting inference, which are mostly due to the use of an improper prior. This problem may motivate the identification of alternative ways to induce sparsity on the spatial maps. 
\newline
\newline
In this work, we modeled the time-shift of each subject as a translation with respect to a common temporal reference. However, since pathological trajectories are different across individuals, it would be valuable to account for individual speed of progressions by introducing a scaling effect, as it has been proposed for example in \cite{ref_koval,ref_schiratti}. This was not in the scope of the current study, as we focused on the analysis of cross-sectional data, thus having only one data point per subject. Therefore, one of the main extensions of this model will be the integration of longitudinal data for each individual, which will allow a more specific time-reparameterization.
\newline
\newline
Our noise model for the reconstruction problem of Equation \ref{eq:data_decomposition} is homoscedastic and i.i.d. Gaussian with zero mean. For this reason, data variability for the entire image is encoded by the variance parameter of the Gaussian noise. Similarly as in standard regression problems, this modelling choice has been motivated to promote simplicity of the model and computational efficiency. However, around $40\%$ of the values in the brain images do not provide relevant information as they represent zero and constant background areas. For this reason, during training, the model can perfectly fit this background and increases its confidence on the overall regression solution, thus lowering the value of the noise variance $\sigma_m$ (cf Figure \ref{fig:vol_biomarkers}). This is in contrast to what we observe with the ADAS13 data (cf Figure \ref{fig:adas13}), where the problem corresponds to standard univariate regression. A potential way to fix this issue could be to train the model only on non-zero image areas, or by implementing an heteroscedastic noise model. However, this latter solution may further increase the number of model parameters. 
\newline
\newline
The modeling results are also sensitive to the specification of the spatio-temporal processes priors. In our case, the monotonicity constraint imposed to the GPs may be too restrictive to completely capture the complexity of the progression of neurodegeneration. From a clinical point of view, the model could also benefit from the integration of data measuring the concentration of Tau protein via PET imaging, in order to quantify key neurobiological processes associated to AD \cite{ref_tau}.
\newline
\newline
In order to guarantee that all the subjects belong to the same pathological trajectory due to AD, the model has only been applied to a cohort of amyloid positive subjects. However, this choice restricts the dynamics of evolution that we could estimate. Indeed, only considering these subjects narrows down the time-line of the pathology, as we study patients at potentially advanced disease stages. Therefore, it would be interesting in a future work to apply the model on a cohort including amyloid negative subjects, to model the brain dynamics over the whole disease natural history. This extension would require to define a proper methodology for disentangling sub-trajectories associated, for example with normal ageing and different pathological subtypes \cite{marco_neurobiology_aging, ref_raphael, ref_alexandra_young}. Moreover, we know that many patients diagnosed with AD can be associated to mixed pathologies such as vascular disease or Lewy bodies. Therefore, a potential clinical application of our method could be to investigate if the spatio-temporal dynamics estimated by MGPA are able to disentangle the contribution of each comorbidity.
\newline
\newline
Assessment of clinical plausibility of MGPA on the ADNI must be corroborated by further validation on independent datasets. Therefore, in a future work, we wish to validate the model on different cohorts to demonstrate its generalization properties. The validation step for each subject would be done by estimating the time-point minimizing the cost between the images of each tested individual, and the image progression model previously estimated on ADNI. The estimated time-shift would provide a measure of the pathological stage of the individual with respect to the modelled trajectory, and could be then compared with the clinical diagnosis of the subject, allowing to test the reliability of our model. This additional validation step could ultimately allow to use the model as a diagnostic instrument of AD. This validation would require an important effort in terms of data harmonisation across multiple cohorts, as well as in terms of clinical interpretation. For this reason, this work will be part of a subsequent publication.
\newline
\newline
We planned to release the source-code along with instructions in order for the model to be used by a large audience. It will be available as a complementary tool on the platform \url{http://gpprogressionmodel.inria.fr/}, which already offers a simple front-end to Gaussian Process Progression model.

\section{Acknowledgements}
This work has been supported by the French government, through the UCA\textsuperscript{JEDI} Investments in the Future project managed by the National Research Agency (ref.n ANR-15-IDEX-01), the grant AAP Santé 06 2017-260 DGA-DSH, and by the Inria Sophia Antipolis - M\'editerran\'ee, "NEF" computation cluster.
\newline
\newline
Data collection and sharing for this project was funded by the Alzheimer's Disease Neuroimaging Initiative (ADNI) and DOD ADNI. ADNI is funded by the National Institute on Aging, the National Institute of Biomedical Imaging and Bioengineering, and through generous contributions from the following: AbbVie, Alzheimer’s Association; Alzheimer’s Drug Discovery Foundation; Araclon Biotech; BioClinica, Inc.; Biogen; Bristol-Myers Squibb Company;CereSpir,  Inc.;Cogstate;Eisai Inc.; Elan Pharmaceuticals, Inc.; Eli Lilly and Company; EuroImmun; F. Hoffmann-La Roche Ltd and its  affiliated  company  Genentech, Inc.;  Fujirebio;  GE  Healthcare; IXICO  Ltd.; Janssen Alzheimer Immunotherapy Research \& Development, LLC.; Johnson \& Johnson Pharmaceutical Research \& Development LLC.;Lumosity;Lundbeck;Merck \& Co., Inc.; Meso Scale Diagnostics, LLC.;NeuroRx Research; Neurotrack Technologies;Novartis Pharmaceuticals Corporation; Pfizer Inc.; Piramal Imaging;Servier; Takeda Pharmaceutical Company; and Transition Therapeutics.The Canadian Institutes of Health Research is providing funds to support ADNI clinical sites in Canada. Private sector contributions are facilitated by the Foundation for the National Institutes of Health (www.fnih.org). The grantee organization is the Northern California Institute for Research and Education, and the study is coordinated by the Alzheimer’s Therapeutic Research Institute at the University of Southern California. ADNI data are disseminated by the  Laboratory  for  Neuro Imaging  at  the University of Southern California. 

\newpage
\bibliographystyle{elsarticle-harv}
\bibliography{tex/bibliography}
\clearpage
\section*{Appendix A.}
\label{sec:appendix_A}
\documentclass[main.tex]{subfiles}
\setcounter{equation}{0}
In this Appendix, we detail the complete derivation of the lower bound.

\begin{align*}
\begin{split}
\log(p(\vect{Y}^{m}, \mathcal{C}^{m}|\vect{Z}^{m},  \vect{\delta}, \sigma_{m}, \gamma_{m})) = \log \Big[ \int & p(\vect{Y}^{m}|\vect{B}^{m},\vect{S}^{m}, \vect{Z}^{m},  \vect{\delta}, \sigma_{m})p(\mathcal{C}|\frac{d\vect{S}^{m}}{d\vect{t}},  \vect{\delta}, \gamma_{m})p(\vect{B}^{m}) \\
& p(\vect{S}^{m},\frac{d\vect{S}^{m}}{d\vect{t}}| \vect{\delta}, \gamma)d\vect{B}^{m}d\vect{S}^{m}\Big] \\
& \hspace{-4.4em} =  \log \Big[ \int p(\vect{Y}^{m}|\vect{B}^{m},\vect{S}^{m}, \vect{Z}^{m},  \vect{\delta}, \sigma_{m})p(\mathcal{C}|\frac{d\vect{S}^{m}}{d\vect{t}},  \vect{\delta}, \gamma_{m})p(\vect{B}^{m}) \\
& p(\frac{d\vect{S}^{m}}{d\vect{t}}|\vect{S}^{m},  \vect{\delta},\gamma)p(\vect{S}^{m})d\vect{B}^{m}d\vect{S}^{m}\Big].
\end{split}
\end{align*}

By observing that $\frac{d\vect{S}^{m}}{d\vect{t}}$ is completely identified by $\vect{S}^{m}$, the equation can be written as:
\begin{align*}
\begin{split}
\log(p(\vect{Y}^{m}, \mathcal{C}^{m}|\vect{Z}^{m},  \vect{\delta},\sigma_{m}, \gamma_{m})) = \log \Big[ \int & p(\vect{Y}^{m}|\vect{B}^{m},\vect{S}^{m}, \vect{Z}^{m},  \vect{\delta}, \sigma_{m})p(\mathcal{C}|\frac{d\vect{S}^{m}}{d\vect{t}},  \vect{\delta}, \gamma_{m})p(\vect{B}^{m}) \\
& p(\vect{S}^{m})d\vect{B}^{m}d\vect{S}^{m}\Big].
\end{split}
\end{align*}

Similarly this derivation can be applied to $\log(p(\vect{V}_{:c}, \mathcal{C}^{c}| \vect{\delta},\nu_{c}, \gamma_{c}))$.

{\allowdisplaybreaks
\begin{align*}
\label{eq:sm1}
\log(p(\vect{Y}^{m}, \mathcal{C}^{m}|\vect{Z}^{m}, \vect{\delta},\sigma_{m}, \gamma_{m})) & = \log \Big[ \int  p(\vect{Y}^{m}|\vect{B}^{m},\vect{S}^{m}, \vect{Z}^{m},  \vect{\delta}, \sigma_{m})p(\mathcal{C}|\frac{d\vect{S}^{m}}{d\vect{t}},  \vect{\delta}, \gamma_{m})p(\vect{B}^{m}) \\
& \hspace{4.5em} p(\vect{S}^{m})d\vect{B}^{m}d\vect{S}^{m}\Big] \\ 
& = \log \Big[ \int  p(\vect{Y}^{m}|\vect{B}^{m},\vect{\Omega}^{m}, \vect{W}^{m}, \vect{Z}^{m}, \vect{\delta}, \sigma_{m})p(\mathcal{C}|\vect{\Omega}^{m}, \vect{W}^{m}, \vect{\delta}, \gamma_{m})p(\vect{B}^{m}) \\
& \hspace{4.5em} p(\vect{\Omega}^{m})p(\vect{W}^{m})d\vect{B}^{m}d\vect{\Omega}^{m}d\vect{W}^{m}\Big]\\
& = \log \Big[ \int  p(\vect{Y}^{m}|\vect{B}^{m},\vect{\Omega}^{m}, \vect{W}^{m}, \vect{Z}^{m}, \vect{\delta}, \sigma_{m})p(\mathcal{C}|\vect{\Omega}^{m}, \vect{W}^{m}, \vect{\delta}, \gamma_{m})p(\vect{B}^{m}) \\
& \hspace{4.5em} p(\vect{\Omega}^{m})p(\vect{W}^{m})\frac{q_1(\vect{B}^{m})q_2(\vect{\Omega}^{m}) q_3(\vect{W}^{m})}{q_1(\vect{B}^{m})q_2(\vect{\Omega}^{m})q_3(\vect{W}^{m})}d\vect{B}^{m}d\vect{\Omega}^{m}d\vect{W}^{m}\Big] \\
& = \log \Big [ \E_{q_1, q_2, q_3} \frac{ p(\vect{Y}^{m}|\vect{B}^{m},\vect{\Omega}^{m}, \vect{W}^{m}, \vect{Z}^{m}, \vect{\delta}, \sigma_{m})p(\mathcal{C}|\vect{\Omega}^{m}, \vect{W}^{m}, \vect{\delta}, \gamma_{m})}{q_1(\vect{B}^{m})q_2(\vect{\Omega}^{m})q_3(\vect{W}^{m})} \\
& \hspace{7em} \frac{p(\vect{B}^{m})p(\vect{\Omega}^{m})p(\vect{W}^{m})}{q_1(\vect{B}^{m})q_2(\vect{\Omega}^{m})q_3(\vect{W}^{m})} \Big] \\
& \geq \E_{q_1, q_2, q_3} \Big( \log \Big [\frac{ p(\vect{Y}^{m}|\vect{B}^{m},\vect{\Omega}^{m}, \vect{W}^{m}, \vect{Z}^{m}, \vect{\delta}, \sigma_{m})p(\mathcal{C}|\vect{\Omega}^{m}, \vect{W}^{m}, \vect{\delta}, \gamma_{m})}{q_1(\vect{B}^{m})q_2(\vect{\Omega}^{m})q_3(\vect{W}^{m})} \\
& \hspace{7.6em} \frac{p(\vect{B}^{m})p(\vect{\Omega}^{m})p(\vect{W}^{m})}{q_1(\vect{B}^{m})q_2(\vect{\Omega}^{m})q_3(\vect{W}^{m})} \Big] \Big) \\
& = \E_{q_1,q_2,q_3}[\log(p(\vect{Y}^{m}|\vect{B}^{m},\vect{\Omega}^{m}, \vect{W}^{m}, \vect{Z}^{m}, \vect{\delta}, \sigma_{m}))] \\
& \hspace{1.2em} + \E_{q_2,q_3}[\log(p(\mathcal{C}^{m}|\vect{\Omega}^{m},\vect{W}^{m}, \vect{\delta}, \gamma_{m}))] \\ 
& \hspace{1.2em} - \mathcal{D}[q_{1}(\vect{B}^{m})||p(\vect{B}^{m})] - \mathcal{D}[q_{2}(\vect{\Omega}^{m})||p(\vect{\Omega}^{m})] - \mathcal{D}[q_{3}(\vect{W}^{m})||p(\vect{W}^{m})].
\end{align*}}

This derivation gives us the lower bound $\vect{\mathcal{L}}_{m}$ of a given modality m. The same technique can be used to derive a lower bound for $\log(p(\vect{V}_{c:},\mathcal{C}^{c}|\vect{\delta},\nu_{c}, \gamma_{c}))$, and by summation over $m$ and $c$  we obtain the lower bound of Equation \ref{eq:total_elbo} for $\log(p(\vect{Y}, \vect{V}, \mathcal{C}|\vect{Z}, \vect{\delta}, \sigma, \nu, \gamma))$.

\section*{Appendix B.}
\label{sec:appendix_B}
\documentclass[main.tex]{subfiles}
\setcounter{equation}{0}
In this section we provide formulas for computing the three KL terms of the lower bound. The total KL divergences are:

\begin{align*}
\begin{split}
& \mathcal{D}[q_{1}(\vect{B})||p(\vect{B})] = \sum_{m} \mathcal{D}[q_{1}(\vect{B}^{m})||p(\vect{B}^{m})], \\
& \mathcal{D}[q_{2}(\vect{\Omega})||p(\vect{\Omega})] = \sum_{m} \mathcal{D}[q_{1}(\vect{\Omega}^{m})||p(\vect{\Omega}^{m})] + \sum_{c} \mathcal{D}[q_{1}(\vect{\Omega}^{c})||p(\vect{\Omega}^{c})], \\
&  \mathcal{D}[q_{3}(\vect{W})||p(\vect{W})] = \sum_{m} \mathcal{D}[q_{3}(\vect{W}^{m})||p(\vect{W}^{m})] + \sum_{c} \mathcal{D}[q_{3}(\vect{W}^{c})||p(\vect{W}^{c})].
\end{split}
\end{align*}

For ease of notation we will drop the $m$ and $c$ indices and will give formulas for a single modality. In \cite{ref_molchanov}, authors provide an approximation of the KL for the maps $\vect{B}$:

\begin{align*}
    -\mathcal{D}[q_{1}(\vect{B})||p(\vect{B})] =  \sum_{n,f} & k_{1}h(k_{2} + k_{3}\log(\alpha_{n,f})) - 0.5\log(1 + \alpha^{-1}_{n,f}) - k_{1},
\end{align*}

where $h$ is the sigmoid function and $k_1 = 0.63576, \ k_2 = 1.87320, \ k_3 = 1.48695$.
\newline

In the case of $\vect{\Omega}$ and $\vect{W}$, we've seen that they have Gaussian priors and approximations which are detailed in Sections \ref{sssec:temporal_sources} and \ref{ssec:vif}. As a result we can obtain closed-form formulas for their KL, leading to:

\begin{align*}
\begin{split}
& \mathcal{D}[q_{2}(\vect{\Omega})|p(\vect{\Omega})] = \frac{1}{2} \displaystyle \sum_{n,j}
\vect{Q}_{n,j}^{2}l_{n} + \vect{R}_{n,j}^{2}l_{n} - 1 - \log(\vect{Q}_{n,j}^{2}l_{n}), \\
& \mathcal{D}[q_{3}(\vect{W})|p(\vect{W})] = \frac{1}{2} \displaystyle \sum_{n,j} \vect{V}_{n,j}^{2} + \vect{T}_{n,j}^{2} - 1 - \log(\vect{V}_{n,j}^{2}).
\end{split}
\end{align*}

By summation over the different modalities we finally obtain the total KL divergences.

\section*{Appendix C.}
\label{sec:appendix_C}
\documentclass[main.tex]{subfiles}
\setcounter{table}{0}
We provide in this Appendix details for the experiments on real data. 

\begin{itemize}
    \item The number of random features for the GP estimation was set to $10$, as it was enough to recover the temporal sources in the synthetic experiments.
    \item The $\gamma$ parameter controlling monotonicity was set to $\gamma_m = 10^{7}$ for each imaging modality ($F_m = 1,418,820$ imaging features and $N_m$ = 6 sources) and $\gamma_c = 1$ for ADAS13 ($C_c =1$ scalar feature).
    \item The lower bound was optimized using the ADAM optimizer \cite{ref_adam}.
    \item We used an alternate optimization scheme between the spatio-temporal parameters and the time-shift of [2000, 1000] iterations repeated 20 times, followed by 30000 iterations in which we only optimized the spatio-temporal parameters.
    \item The expectation terms in the lower bound were approximated using only one Monte-Carlo sample as proposed in \cite{ref_kingma}.
    \item The table below gives the learning rates (LR) of all the parameters of the model.
\end{itemize}

\begin{table*}[h!]
\caption{Learning rates (LR) of the different parameters of the model.}
\label{lr}
\vskip 0.15in
\begin{center}
\begin{small}
\begin{sc}
\begingroup
\setlength{\tabcolsep}{10pt}
\begin{tabular}{ccccccc}
\toprule
 & $\vect{\theta}$ & $\vect{M}$ & $\vect{P}$ & $\vect{Z}$ & $\sigma, \ \nu$ & $\vect{\delta}$  \\
\midrule
lr & $10^{-2}$ & $10^{-3}$ & $10^{-1}$ & $10^{-1}$ & $10^{-2}$ & $10^{-4}$ \\
\bottomrule
\end{tabular}
\endgroup
\end{sc}
\end{small}
\end{center}
\vskip -0.1in
\end{table*}

\clearpage

\section*{Appendix D.}
\label{sec:appendix_D}
\documentclass[main.tex]{subfiles}
In this Appendix, we first provide a pseudo-code for sampling from a normal distribution using the reparameterization trick (see Algorithm \ref{alg:pseudo_code1}). The second pseudo-code (Algorithm \ref{alg:pseudo_code2}) details the steps to compute the lower bound $\vect{\mathcal{L}}_{m}$ for a given imaging modality m. We recall that we want to optimize the following sets of parameters (see Section \ref{ssec:vif}):  $\vect{\delta} = \{\delta_p\}_{p=0}^{P}$, $\vect{Z}$, $\sigma = \{\sigma_m\}_{m=1}^{M}$, $\nu = \{\nu_c\}_{c=1}^{C}$, $\vect{\theta} = \{\theta_m\}_{m=1}^{M} \cup \{\theta_c\}_{c=1}^{C}$, and $\vect{\psi}  = \{\psi_m\}_{m=1}^{M}$. Where $P$ is the number of subjects, $M$ the number of imaging modalities, $C$ the number of scalar features, and $N_{m}$ the number of spatio-temporal sources for a given modality m. 

\begin{align}
\begin{split}
    \vect{\theta} & = \{\vect{R}_{n:}^{m}, \vect{Q}_{n:}^{m}, \vect{T}_{n:}^{m}, \vect{V}_{n:}^{m}, l_{n}, n \in [1, N_{m}]\}_{m=1}^{M} \cup \{\vect{R}_{c:}, \vect{Q}_{c:}, \vect{T}_{c:}, \vect{V}_{c:}, l_{c}, \}_{c=1}^{C}, \\
    \vect{\psi}  & = \{\vect{M}_{n:}^{m}, \vect{P}_{n:}^{m}, n \in [1, N_{m}]\}_{m=1}^{M}.
\end{split}
\end{align}

Similarly to Algorithm \ref{alg:pseudo_code2}, we can derive a function LOSS\_SCALAR when dealing with scalar scores by removing the computations on the spatial sources. Finally the last pseudo-code (Algorithm \ref{alg:pseudo_code3}) details the model optimization. For sake of clarity we denote by $\vect{\Pi}$, the set of all the spatio-temporal parameters of the model.

\begin{algorithm}
\caption{Sampling from $\mathcal{N}(\vect{\mu}, \vect{\Sigma})$ using the reparameterization trick.}
\label{alg:pseudo_code1}
\begin{algorithmic}[1]

\Function{RT}{$\vect{\mu}, \vect{\Sigma}$}

    \State $\vect{\epsilon} \leftarrow \textrm{random sample from} \ \mathcal{N}(\vect{0}, \vect{I})$  
    \State $\vect{z} = \vect{\mu} + \vect{\Sigma}^{\frac{1}{2}}\vect{\epsilon}$ \Comment{Gives one sample from $\mathcal{N}(\vect{\mu}, \vect{\Sigma})$}
    \State \textbf{Return} $\vect{z}$

\EndFunction

\end{algorithmic}
\end{algorithm}

\begin{spacing}{2}
\begin{algorithm}
\caption{Compute loss for a given imaging modality m.}
\label{alg:pseudo_code2}
\begin{algorithmic}[1]

\Function{loss\_image}{$\vect{Y}^{m},\theta_m, \psi_m, \vect{Z}^m, \sigma_m, \vect{\delta}, \gamma_m, N_m, F_m, P$}

\Statex $\textrm{For ease of notation we drop the m index in the pseudo-code.}$

\For{n=1 to N} \Comment{For each source}
\State $\vect{B}_{n:} = \ \textrm{RT}(\vect{M}_{n:}, diag(\vect{P}_{n,:}))$ \Comment{Sampling from $q_1$}

\State $\vect{\omega}^{n} = \ \textrm{RT}(\vect{R}_{n:}, diag(\vect{Q}^{2}_{n:}))$ \Comment{Sampling from $q_2$}

\State $\vect{w}^{n} = \ \textrm{RT}(\vect{T}_{n:}, diag(\vect{V}^{2}_{n:}))$ \Comment{Sampling from $q_3$}

\State $\vect{A}_{n:} = \vect{B}_{n:}\vect{\Sigma}^{n}$  \Comment{Convolution of the sparse code of source n at a given spatial resolution}

\State $\vect{S}_{:n}(\vect{\delta}) = \phi(\vect{\delta}(\vect{\omega}^{n})^{T})\vect{w}^{n}$ \Comment{Compute temporal trajectory of source n}

\State $\vect{S}_{:n}'(\vect{\delta}) = \frac{d\phi(\vect{\delta}(\vect{\omega}^{n})^{T})}{d\vect{\delta}}\vect{w}^{n}$ \Comment{Compute derivative of temporal trajectory of source n}
\EndFor

\State $\vect{\Omega} \leftarrow \textrm{\ block \ diagonal \ matrix  containing all \ the \ set \ of} \ (\vect{\omega}^{n})^{T}$

\State $\vect{W} \leftarrow \textrm{\ block \ diagonal \ matrix  containing \ all \ the \ set \ of} \ \vect{w}^{n}$

\State $\E_{q_1, q_2, q_3}[\log(p(\vect{Y}|\vect{B},\vect{\Omega}, \vect{W}, \vect{Z}, \vect{\delta}, \sigma))] \approx \sum_{p} -\frac{F}{2}\log(2\pi \sigma^{2}) -\frac{1}{2\sigma^{2}} ||\vect{Y}_{p:} - \vect{S}_{p:}\vect{A} - \vect{Z}_{p:}||^{2}$

\State $\E_{q_2,q_3}[\log(p(\mathcal{C}|\vect{\Omega},\vect{W}, \vect{\delta}, \gamma))] \approx - \sum_{p,n} \log((1 + \exp(-\gamma \vect{S}_{p,n}'(\vect{\delta})))$

\Comment{The two expectations terms are approximated using only one Monte-Carlo sample as proposed in \cite{ref_kingma}.}

\State $\textrm{KL} = \mathcal{D}[q_{1}(\vect{B})||p(\vect{B})] + \mathcal{D}[q_{2}(\vect{\Omega})||p(\vect{\Omega})] + \mathcal{D}[q_{3}(\vect{W})||p(\vect{W})]$ \Comment{This tern is computed using approximations and formulas of Appendix B.}

\State $\vect{\mathcal{L}} = \E_{q_1, q_2, q_3}[\log(p(\vect{Y}|\vect{B},\vect{\Omega}, \vect{W}, \vect{Z}, \vect{\delta}, \sigma))] + \E_{q_2,q_3}[\log(p(\mathcal{C}|\vect{\Omega},\vect{W}, \vect{\delta}, \gamma))] - \textrm{KL}$

\State \textbf{Return} $\vect{\mathcal{L}}$

\EndFunction

\end{algorithmic}
\end{algorithm}
\end{spacing}

\begin{spacing}{2}
\begin{algorithm}
\caption{Model optimization.}
\label{alg:pseudo_code3}
\begin{algorithmic}[1]

\Function{optimize}{$\vect{Y}, \vect{V},\vect{\Pi}, \vect{\delta}, \textrm{n\_iter0, \ n\_iter1, \ n\_iter2}$}

\State $\textrm{Initialize} \ \vect{\Pi}^{(0)}, \ \vect{\delta}^{(0)}$
\State $i,j,k = 0$

\While{i $\leq$ n\_iter0}

\For{l=1 to n\_iter1} \Comment{Optimizing spatio-temporal parameters only}

\State $\vect{\mathcal{L}} = 0$

\For{m=1 to M}  \Comment{For each modality}
\State $\vect{\mathcal{L}} \pluseq \textrm{LOSS\_IMAGE}(\vect{Y}^{m}, \theta_m, \psi_m, \vect{Z}^{m}, \sigma_{m}, \vect{\delta}, \gamma_m, N_{m}, F_m, P)$
\EndFor

\For{c=1 to C} \Comment{For each scalar feature}
\State $\vect{\mathcal{L}} \pluseq \textrm{LOSS\_SCALAR}(\vect{V}_{:c}, \theta_c, \nu_{c}, \vect{\delta}, \gamma_m, P)$
\EndFor

\State $\textrm{Compute} \ \frac{d\vect{\mathcal{L}}}{d\vect{\Pi}^{(j)}} \ \textrm{through \ backpropagation}$ 
\State $\vect{\Pi}^{(j+1)} = \textrm{ADAM}(\frac{d\vect{\mathcal{L}}}{d\vect{\Pi}^{(j)}}, \vect{\Pi}^{(j)},
\textrm{LR}(\vect{\Pi}))$  \Comment{The spatio-temporal parameters are optimized by gradient descent using the ADAM optimizer. LR refers to the overall set of learning rates (cf Appendix C.)}

\State $j \pluseq 1$

\EndFor

\For{l=1 to n\_iter2} \Comment{Optimizing time-shift only}

\State $\vect{\mathcal{L}} = 0$

\For{m=1 to M}
\State $\vect{\mathcal{L}} \pluseq \textrm{LOSS\_IMAGE}(\vect{Y}^{m}, \theta_m, \psi_m, \vect{Z}^{m}, \sigma_{m}, \vect{\delta}, \gamma_c, N_{m}, F_m, P)$
\EndFor

\For{c=1 to C}
\State $\vect{\mathcal{L}} \pluseq \textrm{LOSS\_SCALAR}(\vect{V}_{:c}, \theta_c, \nu_{c}, \vect{\delta}, \gamma_c, P)$
\EndFor

\State $\textrm{Compute} \ \frac{d\vect{\mathcal{L}}}{d\vect{\delta}^{(k)}} \ \textrm{through \ backpropagation}$ 

\State $\vect{\delta}^{(k+1)} = \textrm{ADAM}(\frac{d\vect{\mathcal{L}}}{d\vect{\delta}^{(k)}}, \vect{\delta}^{(k)},
\textrm{LR}(\vect{\delta}))$ 

\State $k \pluseq 1$

\EndFor

\State $i \pluseq 1$

\EndWhile

\EndFunction

\end{algorithmic}
\end{algorithm}
\end{spacing}

\clearpage

\section*{Appendix E.}
\label{sec:appendix_E}
\documentclass[main.tex]{subfiles}
\setcounter{table}{0}
\setcounter{figure}{0}

In this Appendix, we show results obtained with standard methods (ICA, NMF, PCA) when applied within the experimental setting of Section \ref{ssec:synthetic_shift}. We recall that for these experiments observations were randomly aligned along the time-axis. The goal was to assess the ability of the different methods to reconstruct the spatio-temporal sources underlying the data when the time-axis is unknown. Results obtained in Table \ref{table_synthetic2} show a substantial decrease of performances for the MSE and SSIM compared to MGPA (cf Table \ref{time_shift_table} in Section \ref{ssec:synthetic_shift}). Indeed, these methods do not consider time as a variable on which inference is required, thus preventing them from reconstructing correctly the temporal sources. Figure \ref{fig:ica_comparison_synthetic2} shows an example of reconstruction when using ICA. We observe that even though the spatial reconstruction remains acceptable, the estimated temporal sources are not interpretable as ICA reconstructs the data using the time-axis on which observations have been mixed.

\begin{table}[h!]
\caption{MSE and SSIM between respectively the ground truth temporal and spatial sources with respect to the ones estimated by the different standard methods.}
\label{table_synthetic2}
\vskip 0.15in
\begin{center}
\begin{small}
\begin{sc}
\begingroup
\setlength{\tabcolsep}{10pt}
\begin{tabular}{ccc}
\toprule
 & Temporal (MSE) & Spatial (SSIM) \\
\midrule
ICA & $ 0.24\pm 0.08$ & $54 \% \pm 2$ \\
NMF & $0.25\pm 0.03$ & $22 \% \pm 14$ \\
PCA & $0.66 \pm 0.05$ & $9 \% \pm 3$  \\
\bottomrule
\end{tabular}
\endgroup
\end{sc}
\end{small}
\end{center}
\vskip -0.1in
\end{table}

\begin{figure}[h!]
\centering
\includegraphics[width=.62\textwidth]{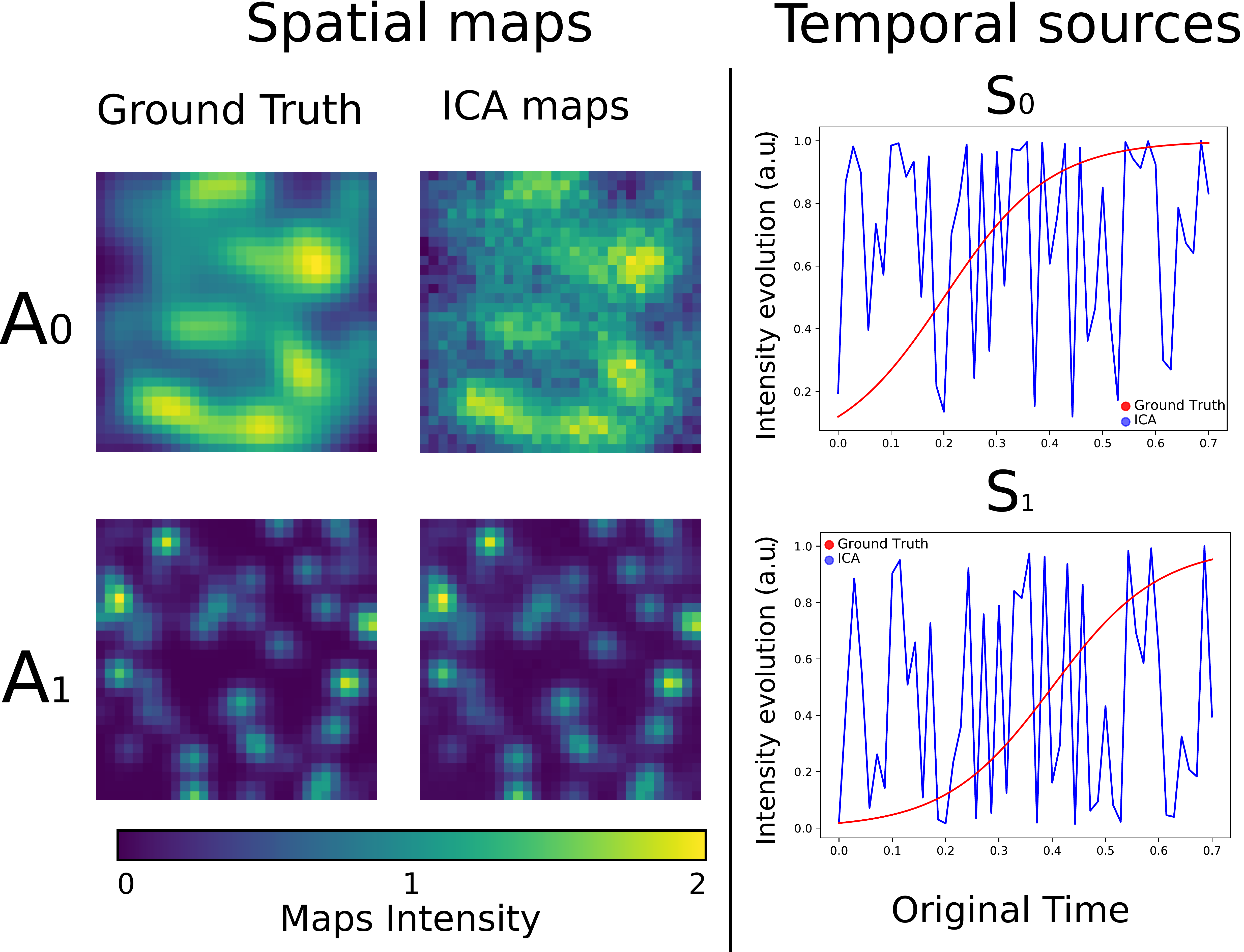}
\caption{Spatial maps: Sample slice from ground truth images ($A_0$ $\lambda = 2$ mm, $A_1$ $\lambda = 1$ mm), the maps estimated by ICA. Temporal sources: Ground truth temporal sources (red) along with sources estimated by ICA (blue).}
\label{fig:ica_comparison_synthetic2}
\end{figure}

\clearpage

\section*{Appendix F.}
\label{sec:appendix_F}
\documentclass[main.tex]{subfiles}
\setcounter{figure}{0}
We provide in this Appendix details on the model convergence when applied on the ADNI data. The training was divided in three iterations of 30000 epochs each. During the two first iterations the spatio-temporal parameters and the time-shift are trained alternatively following a scheme of [2000,1000] epochs ten times. The third iteration only optimizes the spatio-temporal parameters. In Figure \ref{fig:convergence}, we show the evolution of the total loss and the different terms composing it during training. The term reconstruction cost stands for $\sum_{m} \E_{q_1, q_2, q_3}[\log(p(\vect{Y}^{m}|\vect{B}^{m},\vect{\Omega}^{m}, \vect{W}^{m}, \vect{Z}^{m}, \vect{\delta}, \sigma_m))]$, monotonicity cost for $\sum_{m} \E_{q_2,q_3}[\log(p(\mathcal{C}^{m}|\vect{\Omega}^{m},\vect{W}^{m}, \vect{\delta}, \gamma_m))]$ and KL for $\sum_m \mathcal{D}[q_{1}(\vect{B}^{m})||p(\vect{B}^{m})] + \mathcal{D}[q_{2}(\vect{\Omega}^{m})||p(\vect{\Omega}^{m})] + \mathcal{D}[q_{3}(\vect{W}^{m})||p(\vect{W}^{m})]$. We observe that through the first two iterations the reconstruction and monotonicity costs decrease, and become stable during the last iteration. Differently, the KL cost increases during the first iteration as the model is driven by the reconstruction and monotonicity constraints. The KL term decreases during the second iteration, thus regularizing the model, before becoming stable during the third iteration. We also note that the graphs in Figure \ref{fig:convergence} show convergence profiles typical of those obtained with stochastic variational inference schemes, such as with Variational Autoencoders or Bayesian Neural Networks.
\begin{figure}[h!]
\centering
\includegraphics[width=.99\textwidth]{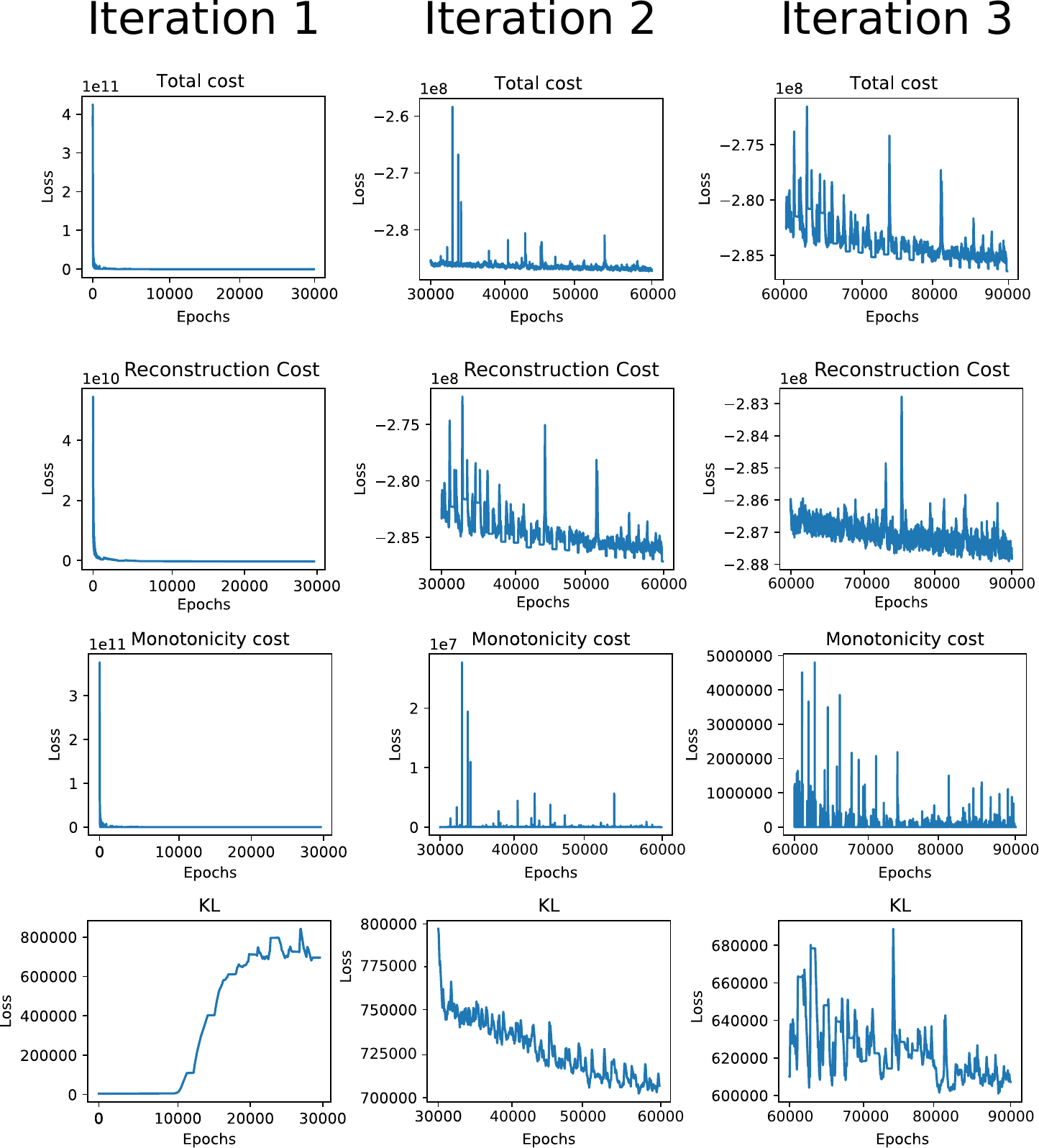}
\caption{Evolution of the total loss, reconstruction cost, monotonicity cost and KL during training. Each iteration corresponds to 30000 epochs.}
\label{fig:convergence}
\end{figure}

\clearpage

\section*{Appendix G.}
\label{sec:appendix_G}
\documentclass[main.tex]{subfiles}
\setcounter{figure}{0}
In this Appendix, we provide the results obtained when applying ICA, NMF and PCA on the ADNI data of Section \ref{sssec:data_preprocessing}. We used the three imaging modalities for each subject and concatenated these images in a $(544 \times 4256460)$ matrix. Our goal was to compare the spatio-temporal processes extracted using these standard methods with the ones from MGPA. We recall that in the case of MGPA the model automatically re-aligns the observations following monotonic assumptions for each biomarker, while these standard methods don't perform any inference on the time variable. Therefore, we created three experimental settings in which we changed the observations' alignment. In the first one, subjects were aligned by their chronological age (Figures \ref{fig:ica_age}, \ref{fig:nmf_age} and \ref{fig:pca_age}), in the second one by ADAS13 (Figures \ref{fig:ica_adas}, \ref{fig:nmf_adas} and \ref{fig:pca_adas}) and in the last one time was randomly initialized like in the experiments of Section \ref{sssec:brain_dynamics} (Figures \ref{fig:ica_no_ordering}, \ref{fig:nmf_no_ordering} and \ref{fig:pca_no_ordering}). We extracted six spatio-temporal sources for each method and each time-alignment, like in \ref{ssec:model_specification}. 
\newline
\newline
We observe that the temporal profiles are generally noisy and hard to interpret due to the lack of constraints on the temporal evolution. This motivates the need of smooth and monotonic constraints as in MGPA. Moreover, due to the concatenation of all the modalities they all share the same temporal patterns. This is an important difference with the modality-specific modelling of MGPA. Finally, we note that the spatial patterns associated with each method are very similar, independently from the time-initialization, while the temporal sources substantially differ. This is also true when time is randomly initialized. These observations point to the challenge of giving a clinical interpretation of the results obtained with these approaches, and therefore to the need of plausible spatio-temporal constraints as provided in MGPA.

\clearpage
\begin{center}
\textbf{\Huge Subjects aligned by age.}\par\medskip
\end{center}
\clearpage

\begin{figure}[h!]
\centering
\includegraphics[width=.95\textwidth]{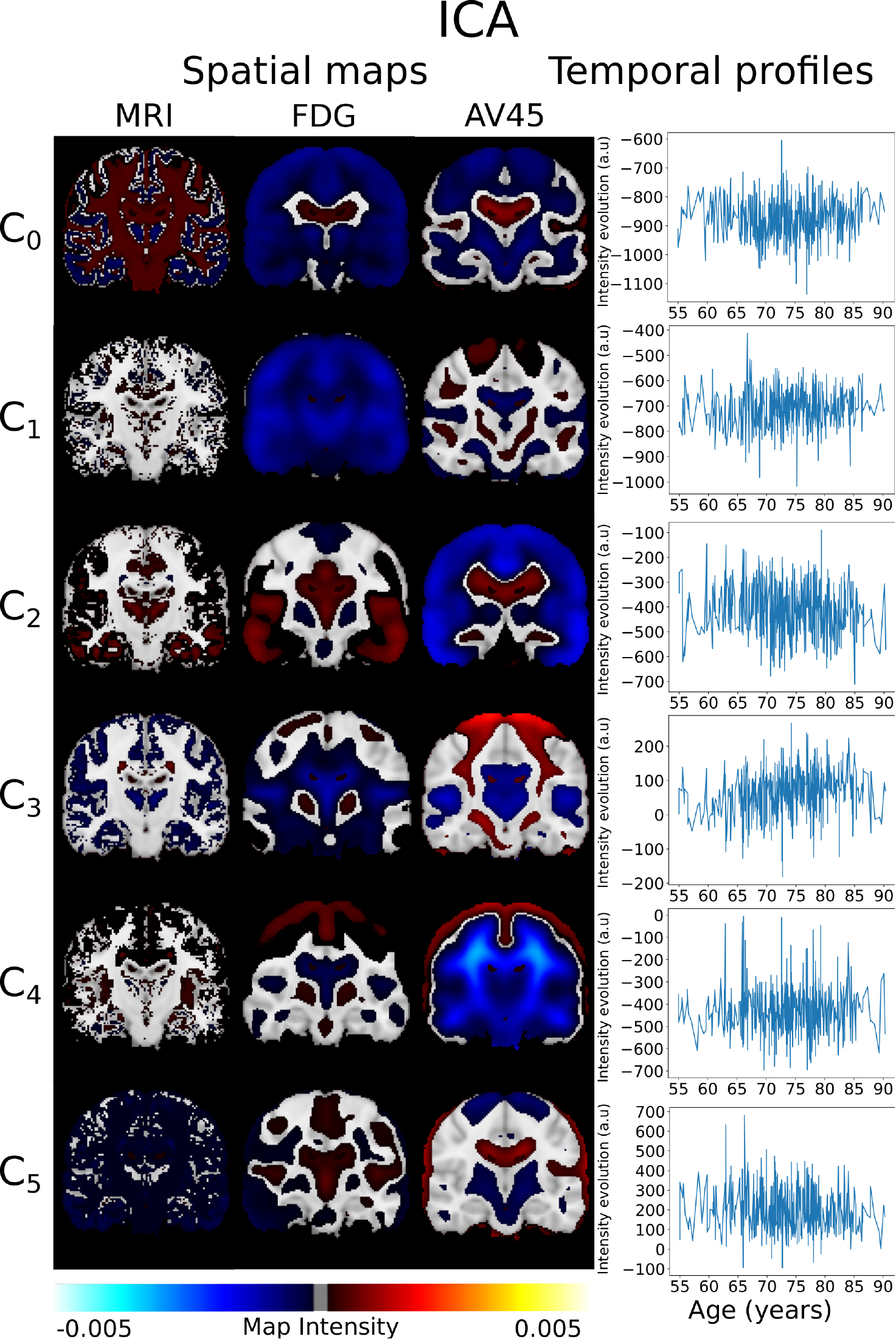}
\caption{Spatio-temporal processes extracted by ICA with subjects aligned by age.}
\label{fig:ica_age}
\end{figure}

\begin{figure}[h!]
\centering
\includegraphics[width=.95\textwidth]{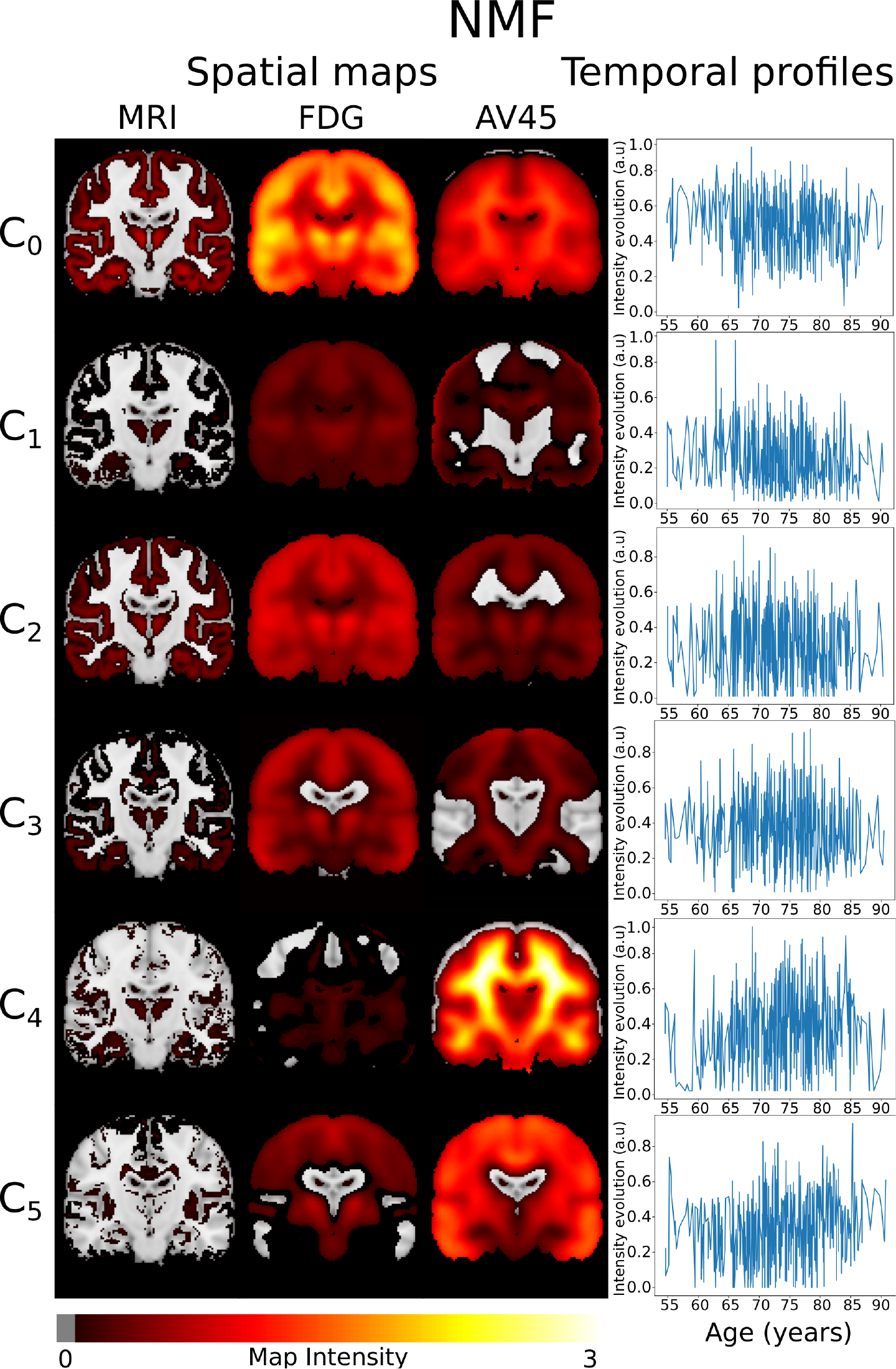}
\caption{Spatio-temporal processes extracted by NMF with subjects aligned by age.}
\label{fig:nmf_age}
\end{figure}

\begin{figure}[h!]
\centering
\includegraphics[width=.95\textwidth]{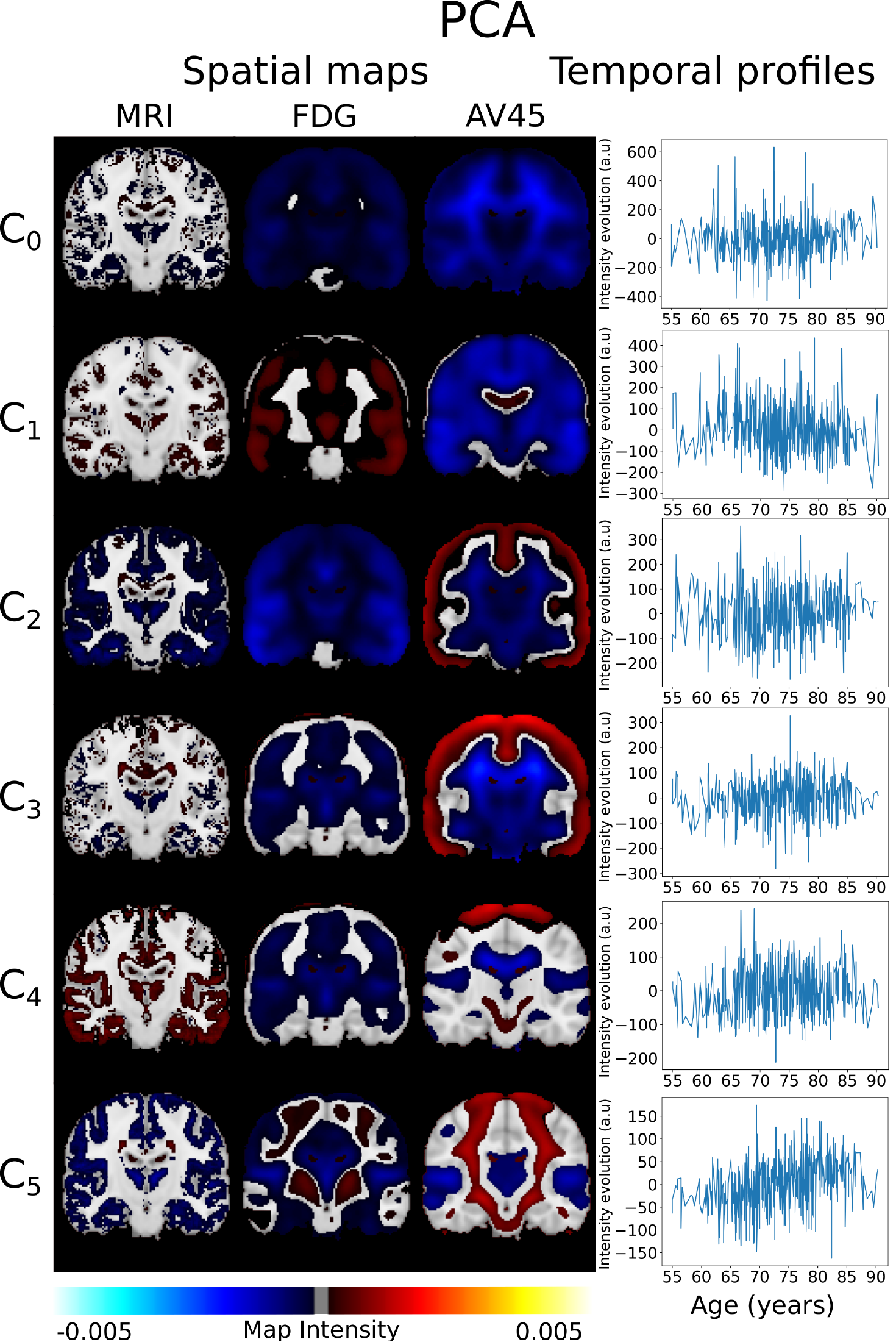}
\caption{Spatio-temporal processes extracted by PCA with subjects aligned by age.}
\label{fig:pca_age}
\end{figure}

\clearpage
\begin{center}
\textbf{\Huge Subjects aligned by ADAS13.}\par\medskip
\end{center}
\clearpage

\begin{figure}[h!]
\centering
\includegraphics[width=.95\textwidth]{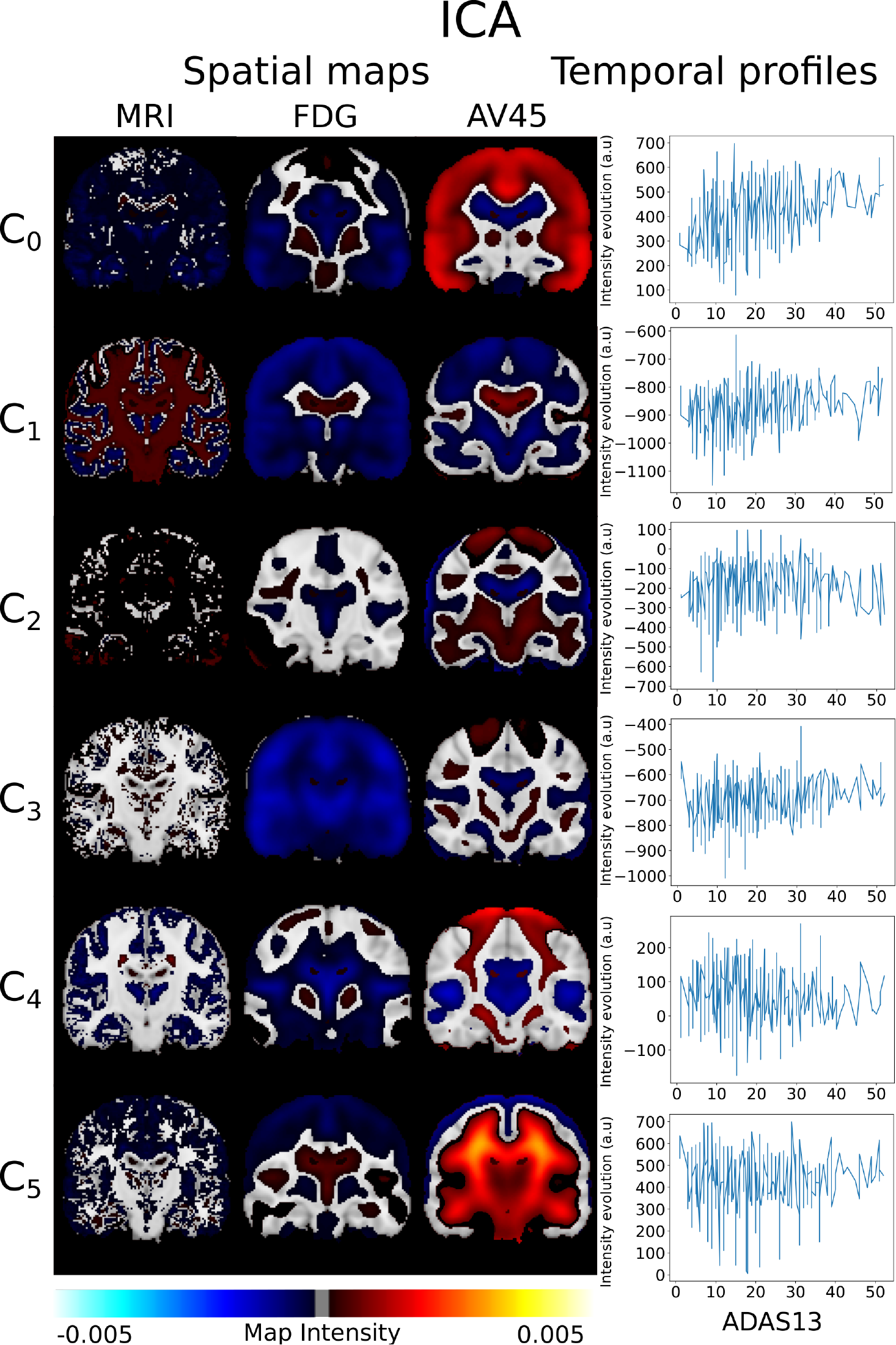}
\caption{Spatio-temporal processes extracted by ICA with subjects aligned by ADAS13.}
\label{fig:ica_adas}
\end{figure}

\begin{figure}[h!]
\centering
\includegraphics[width=.95\textwidth]{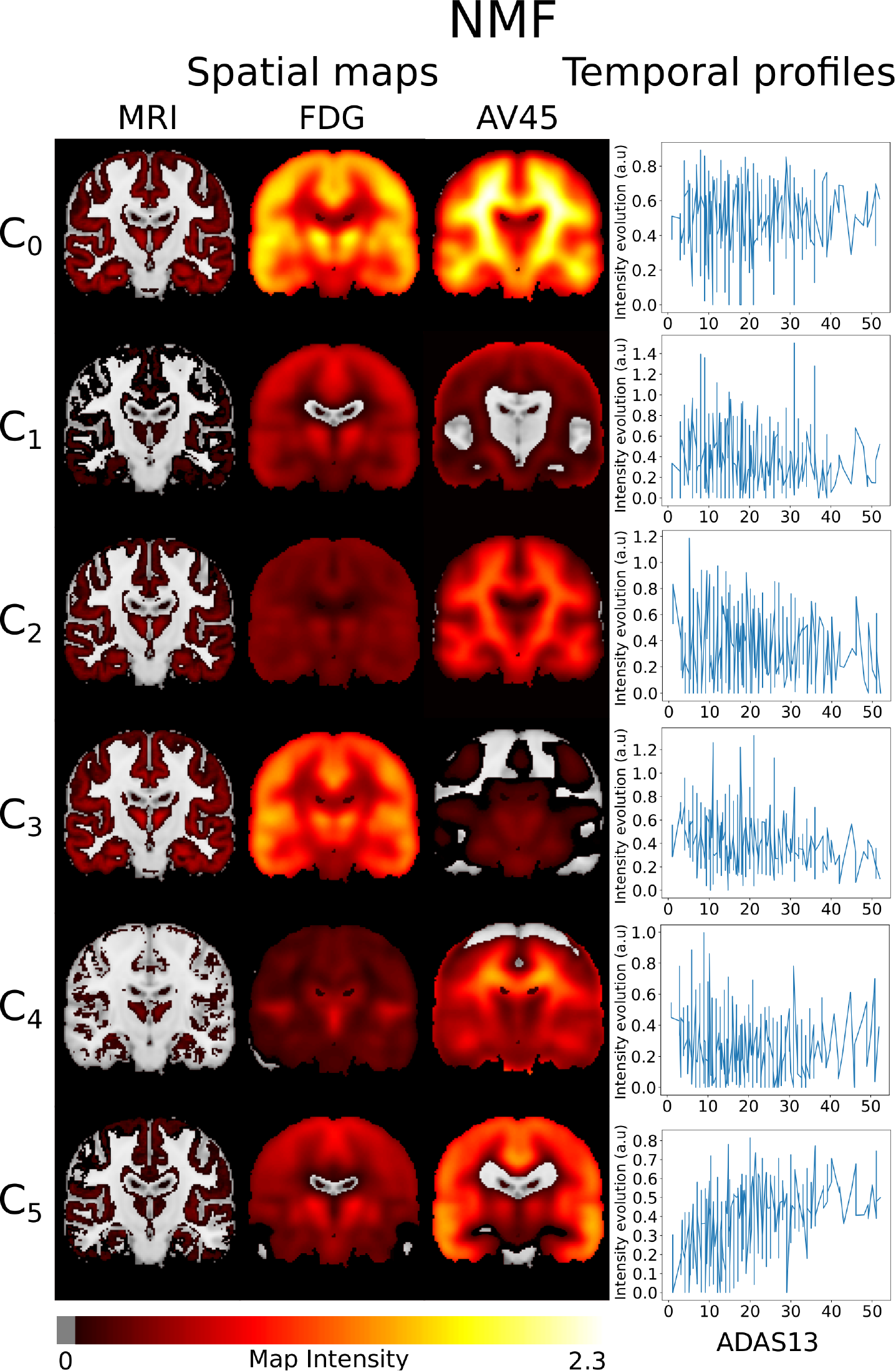}
\caption{Spatio-temporal processes extracted by NMF with subjects aligned by ADAS13.}
\label{fig:nmf_adas}
\end{figure}

\begin{figure}[h!]
\centering
\includegraphics[width=.95\textwidth]{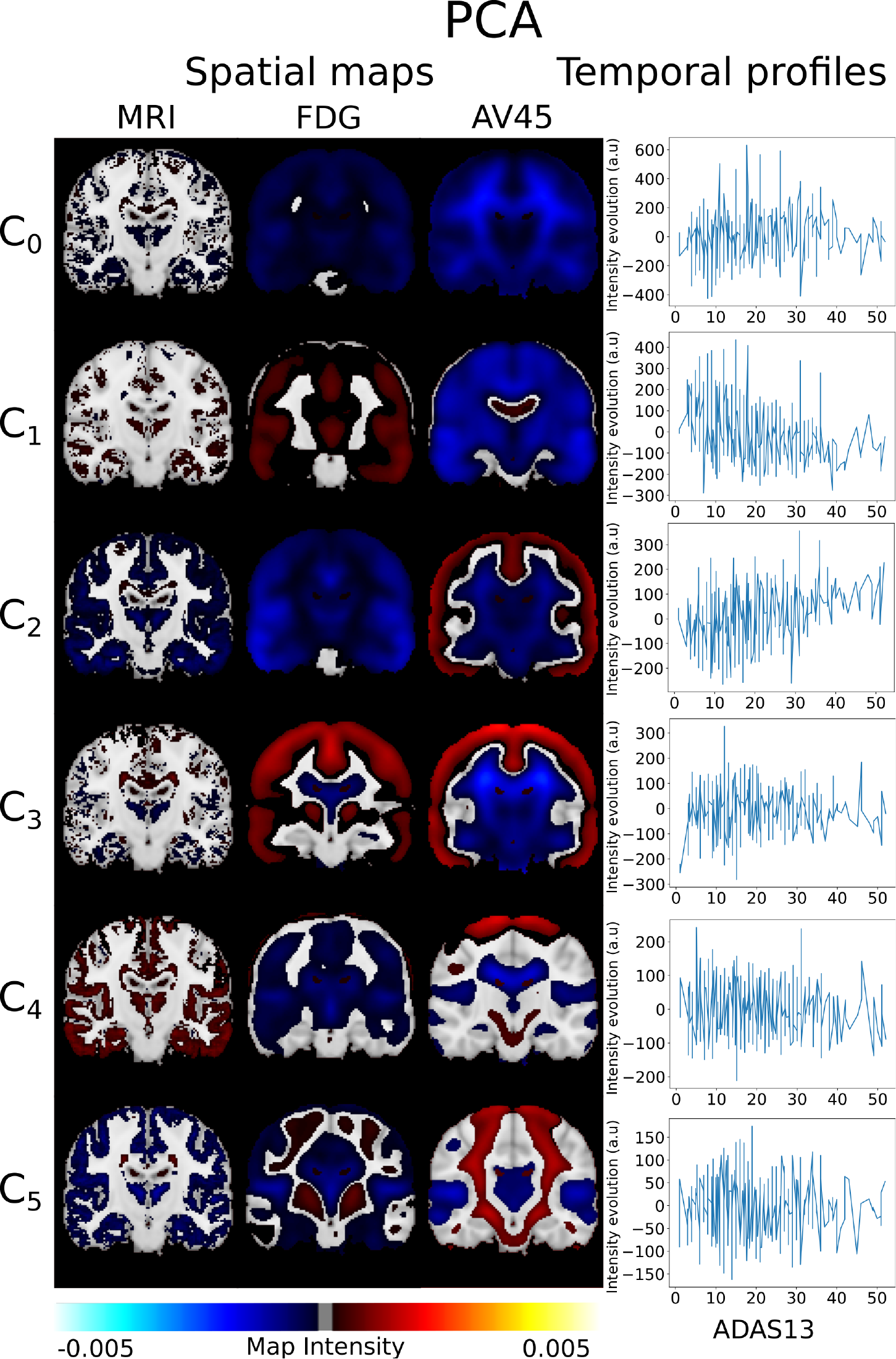}
\caption{Spatio-temporal processes extracted by PCA with subjects aligned by ADAS13.}
\label{fig:pca_adas}
\end{figure}

\clearpage
\begin{center}
\textbf{\Huge Subjects randomly aligned.}\par\medskip
\end{center}
\clearpage

\begin{figure}[h!]
\centering
\includegraphics[width=.95\textwidth]{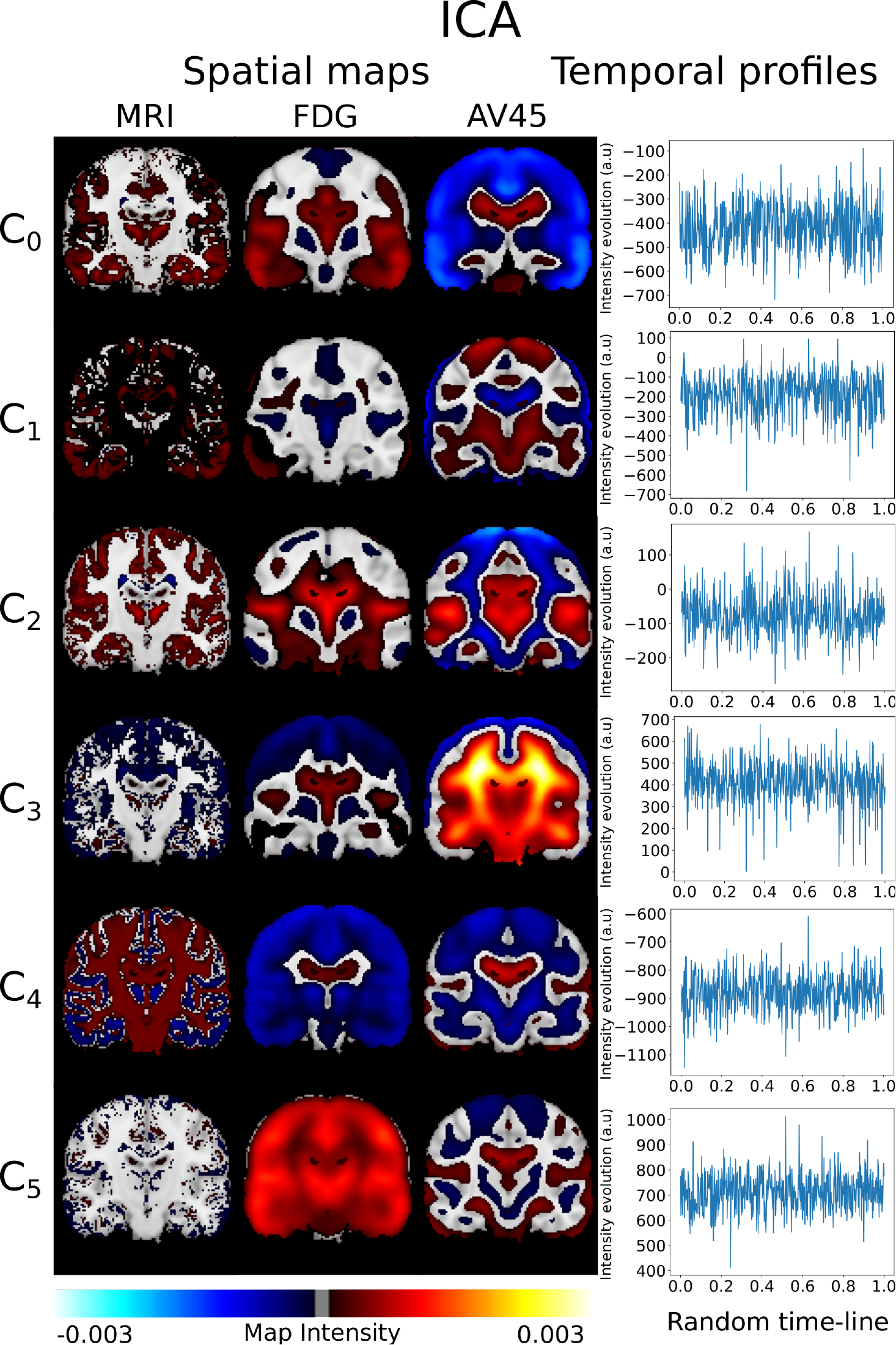}
\caption{Spatio-temporal processes extracted by ICA with subjects randomly aligned.}
\label{fig:ica_no_ordering}
\end{figure}

\begin{figure}[h!]
\centering
\includegraphics[width=.95\textwidth]{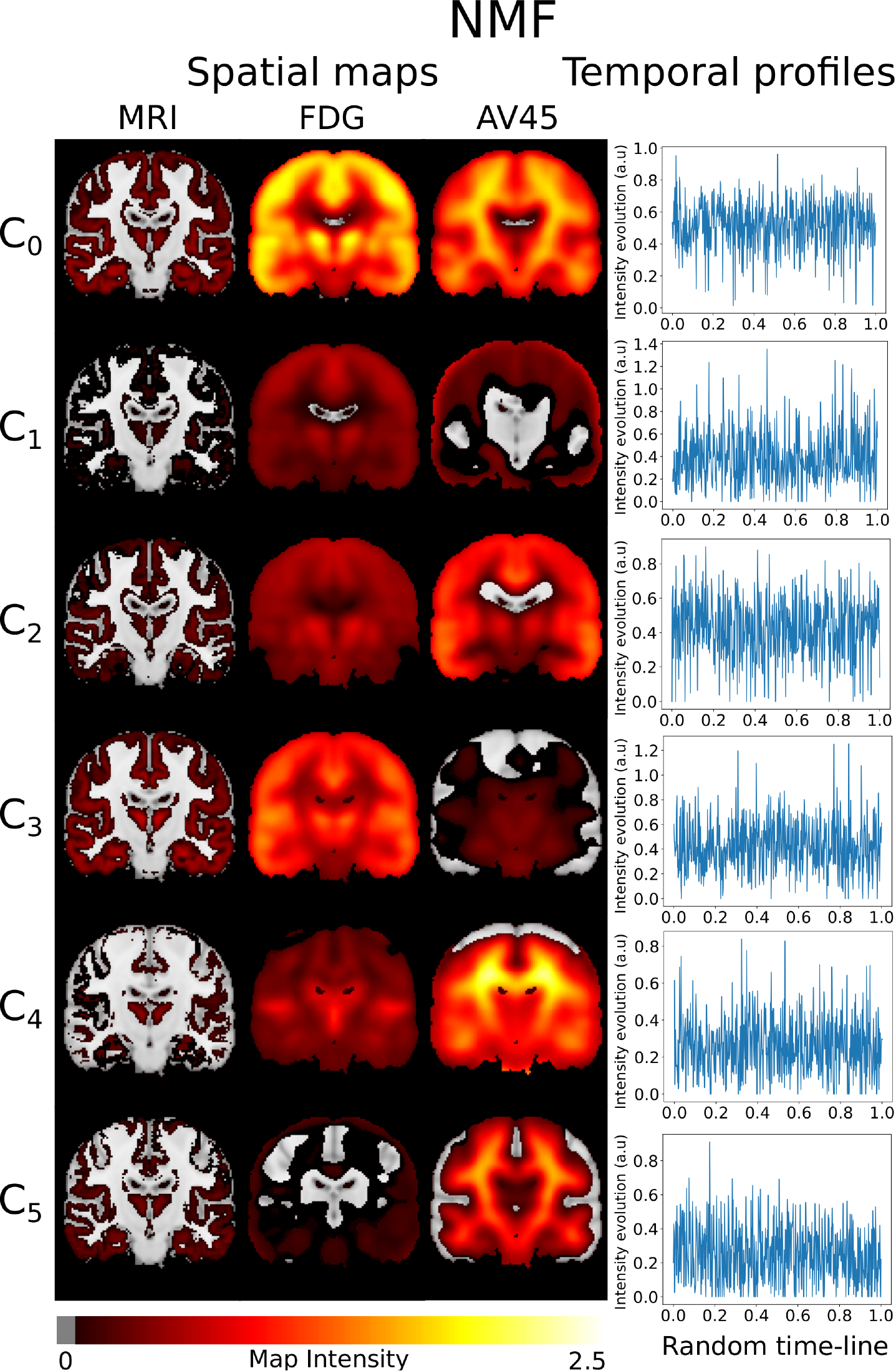}
\caption{Spatio-temporal processes extracted by NMF with subjects randomly aligned.}
\label{fig:nmf_no_ordering}
\end{figure}

\begin{figure}[h!]
\centering
\includegraphics[width=.95\textwidth]{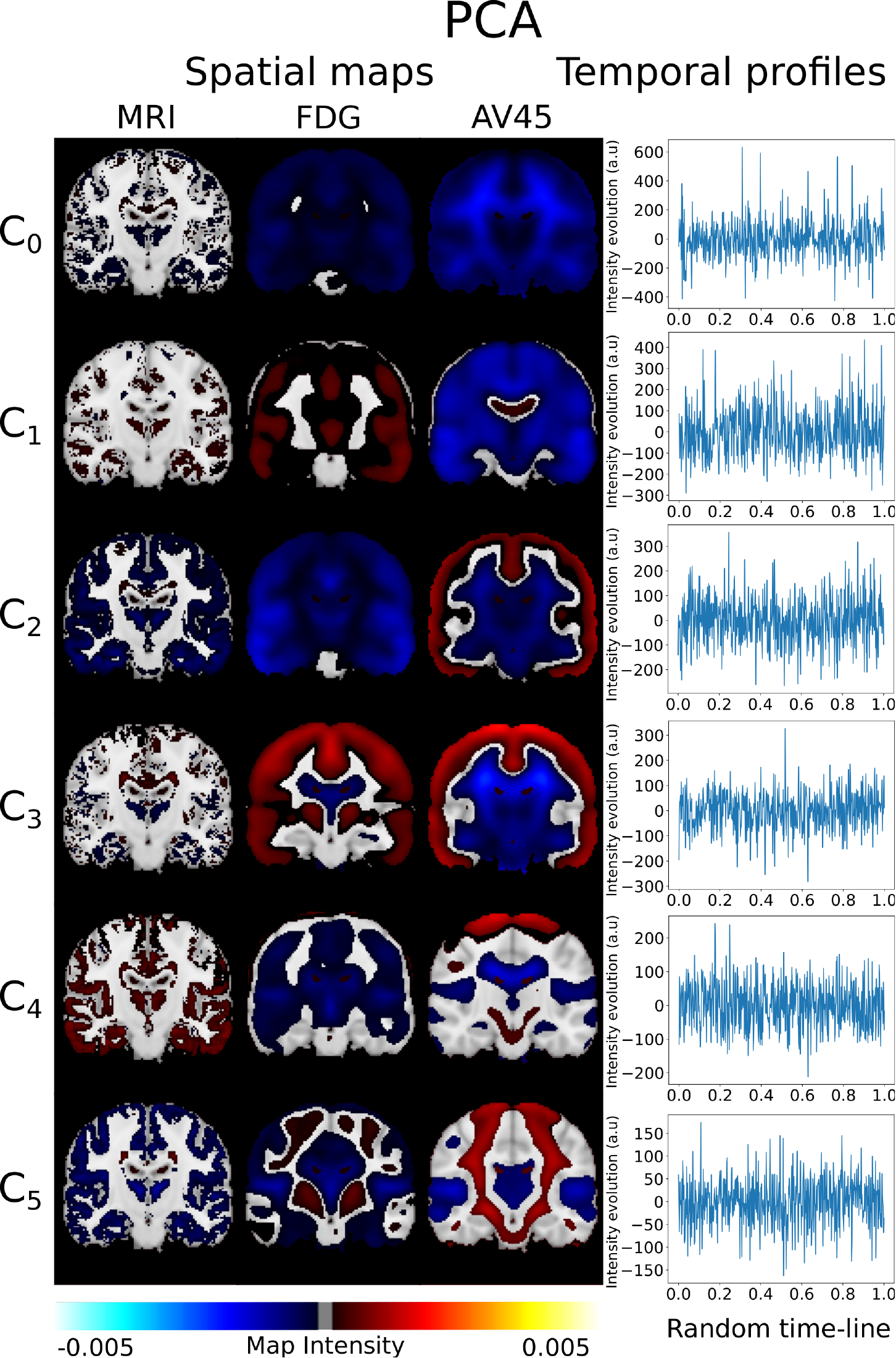}
\caption{Spatio-temporal processes extracted by PCA with subjects randomly aligned.}
\label{fig:pca_no_ordering}
\end{figure}

\end{document}